\documentclass[letterpaper]{article} 
\usepackage{aaai2027}  
\nocopyright
\usepackage[hyphens]{url}  
\usepackage{graphicx} 
\usepackage{natbib}  
\usepackage{caption} 
\usepackage{algorithm}
\usepackage{algorithmic}
\usepackage{amsmath}
\usepackage{adjustbox}
\usepackage[table]{xcolor}
 \usepackage{tabularx}
\usepackage{amssymb}
\usepackage{makecell}
\definecolor{ourrow}{RGB}{235,245,255}
\definecolor{cvprblue}{RGB}{33,113,181}
\definecolor{upperrow}{RGB}{232,242,255}
\usepackage{multirow}
\usepackage{listings}
\DeclareCaptionStyle{ruled}{labelfont=normalfont,labelsep=colon,strut=off} 
\floatstyle{ruled}
\newfloat{listing}{tb}{lst}{}
\floatname{listing}{Listing}

\usepackage{booktabs}

\title{MedPixel: A Unified Pixel-Language Model for Medical Reasoning and Segmentation}
\author{
Haoyu Yang\textsuperscript{\rm 1},
Meixing Shi\textsuperscript{\rm 1},
Zengjie Chen\textsuperscript{\rm 1},
Haoran Sun\textsuperscript{\rm 2}\\
Haitao Leng\textsuperscript{\rm 3},
Xiaoming Shi\textsuperscript{\rm 4},
Yuxiang Cai\textsuperscript{\rm 1},
Yankai Jiang\textsuperscript{\rm 5}
}

\affiliations{
\textsuperscript{\rm 1}Zhejiang University \quad
\textsuperscript{\rm 2}Fudan University \quad
\textsuperscript{\rm 3}Kuaishou\\
\textsuperscript{\rm 4}East China Normal University \quad
\textsuperscript{\rm 5}Shanghai Artificial Intelligence Laboratory\\
\{yanghaoyu, shimeixing, caiyuxiang\}@zju.edu.cn;
jyk1996ver@zju.edu.cn
}

\begin{document}

\maketitle

\begin{abstract}
Reliable medical image understanding requires models to connect clinical language and visual reasoning with pixel-level grounding. Yet medical vision-language models often lack precise localization, whereas medical segmenters typically rely on explicit target categories or precise spatial prompts. This divide is reinforced by a supervision mismatch: segmentation datasets provide precise masks but little language supervision, whereas medical vision-language data rarely pair language with dense spatial annotations. To address this gap, we present \textbf{MedPixel}, a unified medical pixel-language model built around a shared language--mask interface. To provide scalable supervision, we introduce \textbf{MedPLG-440K}, comprising approximately 440K pixel-language task samples constructed through a clinically motivated synthesis process without external LLM annotation. MedPixel is trained with joint multi-task supervised fine-tuning followed by \textbf{Pixel-Level Preference Optimization}, which uses ground-truth masks as offline verifiers to derive response preferences from mask quality. MedPixel supports a broad spectrum of tasks spanning explicit grounding, implicit reasoning, spatial interaction, grounded explanation, and medical VQA. Across this task spectrum, MedPixel achieves strong performance in both pixel-level prediction and response generation, together with effective zero-shot transfer to external grounding benchmarks and robustness to imperfect spatial prompts. Code and model checkpoints will be released at\url{https://github.com/yhy-whu/Medpixel}.
\end{abstract}


\section{Introduction}
\label{sec:introduction}

\begin{figure*}[t]
\centering
\includegraphics[width=0.9\textwidth]{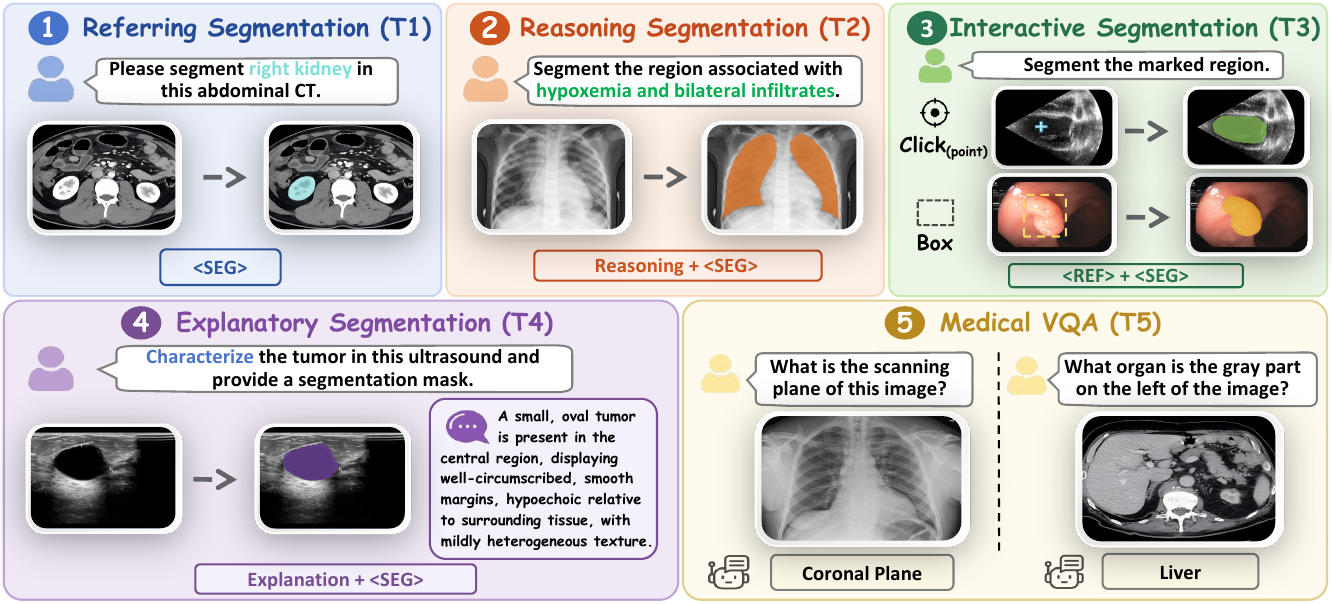}
\caption{Representative task interfaces supported by MedPixel.
T1--T4 produce pixel-grounded outputs through the
\texttt{<SEG>} interface, with T3 additionally incorporating point or
box guidance through \texttt{<REF>}; T5 produces text-only answers.
MedPixel unifies referring, reasoning, interactive, and explanatory segmentation with medical VQA in a single framework.}
\label{fig}
\end{figure*}

Medical image analysis is evolving from isolated recognition and segmentation tasks toward general systems that can support broader clinical decision making. Clinicians synthesize imaging findings with anatomical knowledge, clinical context, and diagnostic intent to identify relevant abnormalities, interpret their significance, and determine where they occur. This evolution expands the role of medical models from processing fixed image inputs to understanding images, language, and spatial prompts, and from producing fixed outputs to supporting reasoning, interaction, localization, and explanation. We summarize these practical requirements into five representative medical task formats: explicit grounding, implicit reasoning, spatial interaction, grounded explanation, and medical visual question answering. Together, these interactions reflect a broader trend toward medical systems that integrate clinical understanding with spatially grounded visual analysis.

Current medical models emphasize different aspects of this functionality. Mainstream medical vision-language models focus primarily on image-level understanding and text generation, including medical question answering, report generation, and clinical reasoning~\cite{llavamed,medgemma,maira2}. Medical segmentation models provide accurate pixel-level predictions across diverse imaging modalities, but usually rely on predefined target categories or explicit spatial prompts~\cite{sam,sam2,medsam,medsam2,sammed2d,biomedparse}.

Recent pixel-language models, including LISA, PixelLM, SAM4MLLM, and UniPixel, connect multimodal language understanding with dense prediction through shared language--mask representations, enabling natural-language grounding, reasoning segmentation, and visual-prompt interaction~\cite{lisa,pixellm,sam4mllm,unipixel}. Related medical models extend pixel-level grounding to biomedical images and increasingly support complex or implicit clinical queries that require target inference before localization~\cite{medplib,unibiomed,medsegr,medreasoner}. Despite this progress, unified support for diverse medical pixel-language interactions remains limited.

As the capability space broadens, supervision becomes a new bottleneck. Different forms of language--pixel interaction often require different types of supervision. This challenge is particularly pronounced in medicine, where existing data are highly asymmetric. Segmentation datasets provide abundant pixel-accurate masks but little language supervision, whereas medical vision-language datasets contain rich questions and answers but rarely pair them with dense spatial annotations. Although recent efforts have expanded biomedical datasets with spatial grounding and reasoning supervision~\cite{unibiomed,medglip,medsegr,medreasoner}, constructing dedicated supervision for each interaction form remains costly. The central challenge is therefore not simply to add more capabilities, but to obtain diverse medical pixel-language supervision in a scalable manner.

This motivates us to reconsider the role of existing segmentation masks. A mask is conventionally treated as the target of a segmentation objective, yet the corresponding annotated region also provides a spatial anchor for connecting image content with language. Together with image, category, and modality information, it can support descriptions of where a finding appears, how it is visually characterized, and what target is being referred to. Inspired by clinical image analysis, these cues can be organized from visual observation and characterization toward target interpretation and spatial grounding. Existing medical masks therefore constitute an underused source of structured pixel-language supervision.

Based on this insight, we construct \textbf{MedPLG-440K} (\textbf{Medical Pixel-Language Grounding 440K}), comprising approximately 440K pixel-language task samples across four grounded interaction formats. These samples are synthesized from existing medical segmentation annotations through a clinically motivated process without external LLM annotation. We separately incorporate medical VQA data to preserve general image-level understanding. Figure~\ref{fig} illustrates the resulting task spectrum.

Built on this supervision, we introduce \textbf{MedPixel}, a unified medical pixel-language model based on Qwen2.5-VL and SAM2~\cite{qwen25vl,sam2}. A shared language--mask interface connects language generation with dense mask decoding through a special \texttt{<SEG>} token, enabling diverse interaction forms to share a common backbone and segmentation pathway.

The shared language--mask interface also creates an alignment problem. The hidden representation of each generated \texttt{<SEG>} token directly conditions mask decoding, yet standard supervised fine-tuning optimizes reference responses and target masks without comparing the spatial consequences of alternative generations. Responses that are similarly plausible in language may therefore produce different \texttt{<SEG>} representations and masks of substantially different quality, creating a mismatch between response likelihood and pixel-level correctness. We introduce \textbf{Pixel-Level Preference Optimization (PLPO)} to address this mismatch. Ground-truth masks act as task-native offline verifiers that rank candidate responses by the quality of their induced masks. These rankings are converted into response preferences, aligning language generation with pixel-level outcomes without requiring an external reward model.

Another practical consideration is the quality of user-provided spatial prompts. Promptable segmentation methods are typically trained and evaluated with precise point or box prompts, whereas real users may provide only approximate spatial guidance. As prompt quality decreases, segmentation performance can degrade substantially~\cite{sam,sam2,medsam,medsam2,roboxsam}. By combining semantic descriptions with spatial cues, MedPixel can use language to clarify the intended target when the spatial prompt is incomplete or inaccurate.

Our contributions are summarized as follows:
\begin{itemize}
\item We introduce \textbf{MedPLG-440K}, comprising approximately 440K pixel-language samples across four grounded interaction formats, synthesized without external LLM annotation.
\item We present \textbf{MedPixel}, a unified model for four pixel-level interactions and medical VQA, together with \textbf{PLPO} for aligning language generation with mask quality.
\item MedPixel achieves leading localization and response-generation
performance, with effective zero-shot transfer and robustness to
imperfect box prompts.
\end{itemize}

\section{Related Work}
\label{sec:related}

\subsection{Medical Vision-Language and Segmentation Models}
Medical vision-language models have advanced medical image understanding through multimodal pretraining and instruction tuning. Representative models such as LLaVA-Med, MedGemma, and MAIRA-2 support medical question answering, report generation, and clinical reasoning~\cite{llavamed,medgemma,maira2}. In parallel, foundation segmentation models such as SAM and SAM2 have inspired medical adaptations including MedSAM, MedSAM2, and SAM-Med2D, while BiomedParse extends category-driven segmentation across diverse biomedical modalities~\cite{sam,sam2,medsam,medsam2,sammed2d,biomedparse}. These two directions provide complementary strengths: vision-language models offer flexible semantic interaction, whereas segmentation models provide precise spatial outputs. This complementary development has motivated models that connect language understanding with pixel-level grounding.

\subsection{Pixel-Level Grounding and Medical Reasoning}
Pixel-level vision-language models connect language understanding and dense prediction. LISA introduces a special segmentation token that maps language-model representations to masks, enabling reasoning segmentation from natural-language instructions~\cite{lisa}. PixelLM extends this paradigm to multi-object grounding, SAM4MLLM incorporates visual prompts for richer multimodal interaction, and UniPixel unifies referring, mask generation, visual prompting, and object-centric reasoning~\cite{pixellm,sam4mllm,unipixel}. Together, these methods make pixel-level prediction an integral part of multimodal interaction and reasoning rather than merely a final output.

This direction has also been extended to medical imaging. MedPLIB and UniBiomed combine medical language understanding with pixel-level prediction across diverse tasks and imaging settings~\cite{medplib,unibiomed}. MedSeg-R reasons over complex clinical instructions beyond explicit target descriptions, while MedReasoner further addresses implicit queries that require target inference before localization~\cite{medsegr,medreasoner}. Building on this progression, MedPixel unifies multiple grounded medical interaction formats, while MedPLG-440K systematically derives their supervision from existing segmentation annotations rather than constructing separate supervision for each format.

\section{Method}
\label{sec:method}

\begin{figure*}[!t]
\centering
\includegraphics[width=\textwidth]{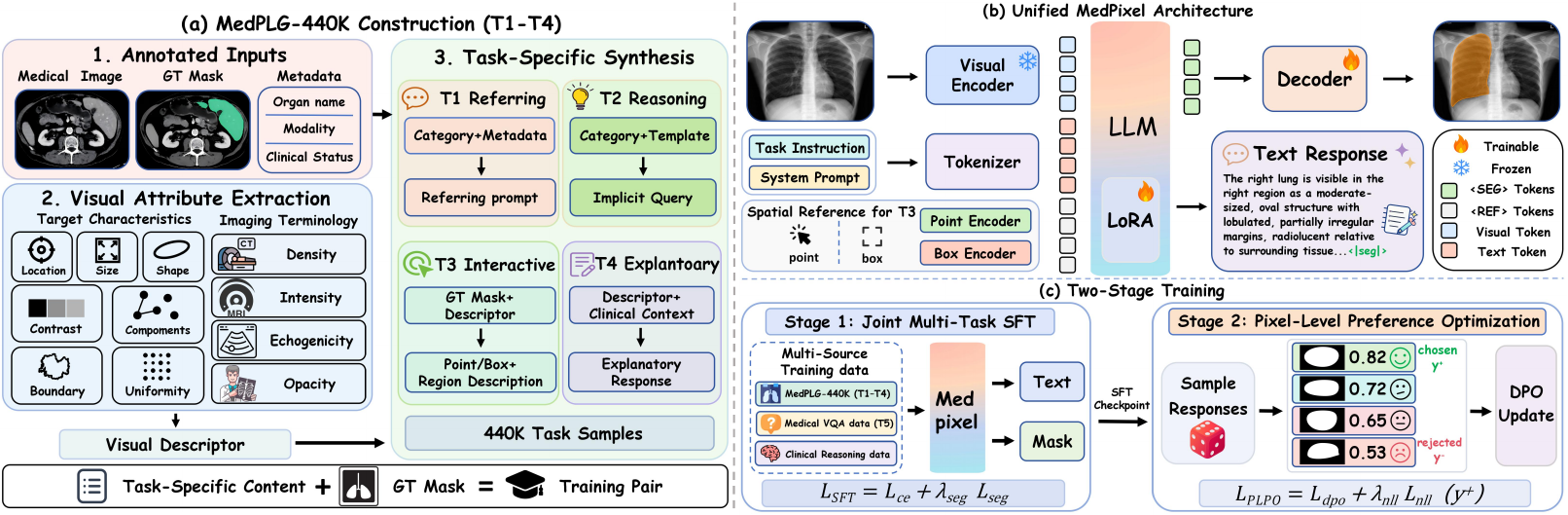}
\caption{Overview of MedPixel.
(a) Construction of MedPLG-440K.
(b) MedPixel architecture for unified language generation and mask prediction.
(c) Two-stage training with multi-task SFT and PLPO.}
\label{fig_fra}
\end{figure*}

Figure~\ref{fig_fra} summarizes the overall design of MedPixel. The framework consists of three parts: (a) MedPLG-440K construction, (b) unified MedPixel architecture, (c) two-stage training strategy. We first build MedPLG-440K from existing medical segmentation annotations, then use it to support a shared language--mask interface for unified text generation and dense prediction, and finally optimize the model with joint multi-task supervised fine-tuning followed by PLPO.

\subsection{Task Interfaces and Architecture}
\label{sec:arch}

\noindent\textbf{Task Formulation.}
Given a medical image $I$, a task instruction $q$, and, for interactive segmentation, an optional spatial reference $r$, MedPixel generates a textual response $y$ and, when required, a segmentation mask $\hat{M}$. As summarized in Figure~\ref{fig}, the first four tasks produce mask-grounded responses, whereas medical VQA produces text-only answers.




\noindent\textbf{Language--Mask Interface.}
MedPixel couples Qwen2.5-VL~\cite{qwen25vl} with a SAM2~\cite{sam2} mask decoding branch. Qwen2.5-VL performs multimodal understanding and language generation, while the mask branch produces dense segmentation outputs. All interaction formats are processed through the same multimodal backbone. Segmentation-oriented responses contain a special \texttt{<SEG>} token, while interactive inputs additionally contain a spatial reference token \texttt{<REF>}.

For segmentation-oriented tasks, MedPixel generates a special \texttt{<SEG>} token at the position where mask prediction is required. Let $h_{\texttt{<SEG>}}$ denote the hidden state of this token. A learnable projection $\Phi_{\text{seg}}$ maps it into the SAM2 prompt embedding space:
\begin{equation}
z_{\text{seg}}
=
\Phi_{\text{seg}}
\left(
h_{\texttt{<SEG>}}
\right).
\end{equation}
Given the SAM2 image embedding $F_I$, the mask decoder predicts
\begin{equation}
\hat{M}
=
\mathcal{D}_{\text{SAM2}}
\left(
F_I,
z_{\text{seg}}
\right),
\end{equation}
where $\mathcal{D}_{\text{SAM2}}$ denotes the mask decoding pathway. This shared language--mask interface allows different query and response formats to use a common dense prediction mechanism.

\noindent\textbf{Spatial Reference Injection.}
Interactive segmentation incorporates point or box guidance through a special \texttt{<REF>} token. Given a spatial reference $r$, a trainable prompt encoder converts it into a sparse spatial representation, which is then projected to the language-model hidden dimension:
\begin{equation}
z_{\text{ref}}
=
\Phi_{\text{ref}}
\left(
\mathcal{E}_{\text{ref}}(r)
\right),
\end{equation}
where $\mathcal{E}_{\text{ref}}$ denotes the spatial prompt encoder and $\Phi_{\text{ref}}$ the projection into the Qwen2.5-VL embedding space. Before the sequence is processed by the language model, $z_{\text{ref}}$ replaces the original token embedding at the \texttt{<REF>} position. Spatial guidance therefore enters the same multimodal sequence as image and text inputs, allowing the subsequent \texttt{<SEG>} representation to integrate semantic and spatial information before mask decoding.

\subsection{MedPLG-440K Construction}
\label{sec:data}

As illustrated in Figure~\ref{fig_fra}(a), MedPLG-440K is constructed from existing medical segmentation annotations. Each source sample provides a medical image $I$, a ground-truth mask $M$, a target category $c$, and metadata $m$. Based on these annotations, we first perform visual attribute extraction to obtain a visual descriptor $d$, and then synthesize four segmentation-oriented interaction formats:
\begin{equation}
(I,M,c,m)
\xrightarrow{\mathcal{A}}
d
\xrightarrow{\mathcal{T}}
\left\{
(x_t,y_t,M)
\right\}_{t=1}^{4},
\end{equation}
where $\mathcal{A}$ denotes visual attribute extraction and $\mathcal{T}$ denotes task-specific synthesis.

\noindent\textbf{Visual Attribute Extraction.}
For each annotated region, we derive a structured visual descriptor from mask geometry and image statistics. The descriptor captures location, size, shape, contrast, component structure, boundary, and uniformity. Location and size are obtained from the mask centroid and relative area; shape and component structure are characterized by region geometry and connected components; boundary regularity is estimated from contour measurements. Contrast and uniformity are computed from differences between the foreground and background and from variation within the region. Appearance attributes are expressed using terminology appropriate to the imaging modality. For example, appearance characteristics may be described in terms of density for CT, signal intensity for MRI, echogenicity for ultrasound, or opacity for X-ray. This process yields a compact visual descriptor aligned with the annotated region.

\noindent\textbf{Task-Specific Synthesis.}
The source annotations and visual descriptor are combined according to the target interaction format. For \textit{T1 referring segmentation}, the target category and metadata are used to construct explicit referring prompts. For \textit{T2 reasoning segmentation}, category information is paired with curated templates to form implicit queries that require target inference before localization. For \textit{T3 interactive segmentation}, points or bounding boxes are sampled
from the ground-truth mask, while the visual descriptor is converted
into a concise region description. For \textit{T4 explanatory segmentation}, the visual descriptor is combined with concise clinical context to construct an explanatory response grounded in the target region. Each synthesized sample is paired with the original ground-truth mask to form a pixel-language training pair.

Through this pipeline, we construct \textbf{MedPLG-440K}, comprising approximately 440K pixel-language task samples. During Stage~1, MedPLG-440K is further combined with medical VQA and auxiliary clinical reasoning data to support joint training across five tasks. Dataset sources and construction details are provided in Appendices A and B.

\subsection{Two-Stage Training}
\label{sec:training}

Figure~\ref{fig_fra}(c) summarizes the two-stage optimization procedure of MedPixel. Stage~1 learns the shared language--mask interface through joint multi-task supervised fine-tuning. Stage~2 further aligns response generation with mask quality through PLPO.

\noindent\textbf{Stage~1: Joint Multi-Task SFT.}
Stage~1 jointly trains MedPixel on a unified mixture of all tasks. The objective combines autoregressive language modeling with mask supervision:
\begin{equation}
\mathcal{L}_{\text{SFT}}
=
\mathcal{L}_{\text{ce}}
+
\lambda_{\text{seg}}\mathcal{L}_{\text{seg}},
\end{equation}
where $\mathcal{L}_{\text{ce}}$ denotes the language modeling loss and $\mathcal{L}_{\text{seg}}$ denotes the pixel-level segmentation loss. For samples with segmentation supervision, the mask objective is
\begin{equation}
\mathcal{L}_{\text{seg}}
=
\lambda_{\text{focal}}\mathcal{L}_{\text{focal}}
+
\lambda_{\text{dice}}\mathcal{L}_{\text{dice}}
+
\lambda_{\text{iou}}\mathcal{L}_{\text{iou}}
+
\lambda_{\text{obj}}\mathcal{L}_{\text{obj}}.
\end{equation}
For samples without mask supervision, the mask term is omitted. This joint objective allows the five task formats to share the same language--mask interface while learning both language generation and pixel-level prediction.

\noindent\textbf{Stage~2: Pixel-Level Preference Optimization.}

Building on the Stage~1 checkpoint, we construct response preference pairs from reasoning segmentation samples according to the
quality of the masks induced by alternative generations. For each input
$x_i$, we sample $N$ candidate responses from $\pi_{\mathrm{SFT}}$ and
decode the mask associated with each generated \texttt{<SEG>}
representation. The quality of the $n$-th candidate is measured against the corresponding ground-truth mask:
\begin{equation}
s_i^{(n)}
=
\operatorname{Dice}
\left(
\hat{M}_i^{(n)},
M_i^{\mathrm{gt}}
\right),
\end{equation}
where $\hat{M}_i^{(n)}$ denotes the mask induced by the $n$-th candidate response. Ground-truth masks are used only for offline scoring and are never exposed during candidate generation. This process converts pixel-level outcomes into response preference supervision without requiring an additional reward model or online rollouts.

We discard empty responses and candidates that do not produce a valid \texttt{<SEG>} output from which a mask can be decoded. Among the remaining candidates, the highest-scoring response is selected as the chosen response $y_i^{+}$, while a lower-scoring valid response is selected as the rejected response $y_i^{-}$. To avoid weak or ambiguous preference supervision, we retain a pair only when
\begin{equation}
s_i^{+}
\geq
\tau_{\min},
\qquad
s_i^{+}
-
s_i^{-}
\geq
\tau_{\mathrm{gap}},
\end{equation}
where $\tau_{\min}$ ensures sufficient quality of the chosen response and $\tau_{\mathrm{gap}}$ enforces a meaningful quality difference between the chosen and rejected responses. Inputs that do not satisfy these criteria are discarded. The resulting preference dataset is
\begin{equation}
\mathcal{P}
=
\left\{
(x_i,y_i^{+},y_i^{-})
\right\}_{i=1}^{|\mathcal{P}|}.
\end{equation}

Given $\mathcal{P}$, we optimize MedPixel using Direct Preference Optimization~\cite{dpo}. Let $\pi_{\theta}$ denote the trainable policy and $\pi_{\mathrm{ref}}=\pi_{\mathrm{SFT}}$ the frozen reference policy. We define the relative preference log-ratio as
\begin{equation}
\Delta_{\theta}
=
\log
\frac{
\pi_{\theta}(y^{+}\mid x)
}{
\pi_{\mathrm{ref}}(y^{+}\mid x)
}
-
\log
\frac{
\pi_{\theta}(y^{-}\mid x)
}{
\pi_{\mathrm{ref}}(y^{-}\mid x)
}.
\end{equation}
The preference objective is
\begin{equation}
\mathcal{L}_{\text{PLPO}}
=
-
\mathbb{E}_{\mathcal{P}}
\left[
\log
\sigma
\left(
\beta \Delta_{\theta}
\right)
\right]
+
\lambda_{\mathrm{nll}}
\mathcal{L}_{\mathrm{nll}}
\left(
y^{+}\mid x
\right),
\end{equation}
where $\beta$ controls the preference strength and the NLL term anchors the model to the chosen response.

During Stage~2, the segmentation projector $\Phi_{\mathrm{seg}}$ and the SAM2 mask decoder remain frozen. Preference optimization therefore reshapes the response distribution and the resulting \texttt{<SEG>} representations while preserving the dense decoding pathway learned in Stage~1. In this way, MedPixel converts differences in pixel-level mask quality into response-level preferences, encouraging generations whose \texttt{<SEG>} representations induce more accurate segmentation outcomes.

\begin{table*}[t]
\centering
\small
\begin{adjustbox}{max width=\textwidth}
\begin{tabular}{@{}l c *{13}{r}@{}}
\toprule
\multirow{2}{*}{\textbf{Method}} & \multirow{2}{*}{\textbf{Size}} & \multicolumn{2}{c}{\textbf{T1} Refer.} & \multicolumn{4}{c}{\textbf{T2} Reason.} & \multicolumn{2}{c}{\textbf{T3} Interact.} & \multicolumn{3}{c}{\textbf{T4} Explan.} & \multicolumn{2}{c}{\textbf{T5} VQA} \\
\cmidrule(lr){3-4}
\cmidrule(lr){5-8}
\cmidrule(lr){9-10}
\cmidrule(lr){11-13}
\cmidrule(l){14-15}
& & Dice & NSD & Dice & NSD & MET. & TGA & Dice & NSD & Dice & NSD & MET. & IMCQ & TMCQ \\
\midrule

\multicolumn{15}{@{}l}{\textit{Segmentation specialists}} \\

MedSAM & 93M & -- & -- & -- & -- & -- & -- & 43.6 & 32.9 & -- & -- & -- & -- & -- \\
MedSAM2 & 38M & -- & -- & -- & -- & -- & -- & 75.7 & 52.4 & -- & -- & -- & -- & -- \\
SAM2 & 0.2B & -- & -- & -- & -- & -- & -- & 61.7 & 31.9 & -- & -- & -- & -- & -- \\
SAM-Med2D & 0.3B & -- & -- & -- & -- & -- & -- & 58.6 & 30.2 & -- & -- & -- & -- & -- \\
BiomedParse & 0.3B & 80.0 & 55.0 & -- & -- & -- & -- & -- & -- & -- & -- & -- & -- & -- \\

\midrule
\multicolumn{15}{@{}l}{\textit{Pixel-language and VQA models}} \\

UniBiomed & 1B & 65.0 & 27.7 & 28.3 & 14.4 & 4.9 & 27.3 & -- & -- & 35.8 & 19.1 & 7.3 & 7.0 & 6.6 \\
Citrus-V & 8B & 45.7 & 27.9 & 37.5 & 25.9 & -- & -- & -- & -- & 35.1 & 23.0 & 10.2 & 41.1 & 51.1 \\
VisionReasoner & 7B & 42.4 & 14.6 & 33.4 & 14.5 & 29.7 & 39.8 & -- & -- & 31.1 & 13.0 & 17.2 & 41.6 & 47.0 \\
LISA++ & 7B & 34.7 & 7.8 & 22.8 & 8.9 & 7.9 & 0.1 & -- & -- & 12.1 & 5.3 & 12.0 & -- & -- \\
SAM4MLLM & 8B & 27.1 & 3.5 & 20.8 & 3.6 & -- & -- & -- & -- & 21.4 & 4.2 & -- & -- & -- \\
LISA & 7B & 26.9 & 7.7 & 26.3 & 13.2 & -- & -- & -- & -- & 23.0 & 9.0 & -- & -- & -- \\
MMedAgent & 7B & 27.8 & 7.9 & 25.6 & 12.8 & -- & -- & -- & -- & 25.4 & 12.5 & -- & 6.7 & 8.8 \\
MedPLIB & 7B & 17.1 & 4.4 & 8.3 & 3.1 & -- & -- & -- & -- & 11.1 & 3.8 & -- & 36.4 & 31.0 \\
PixelLM & 7B & 30.6 & 3.3 & 18.2 & 3.0 & 8.1 & 4.9 & -- & -- & 10.0 & 1.5 & 11.6 & -- & -- \\
UniPixel & 7B & 39.0 & 9.6 & 27.9 & 7.5 & 3.0 & 0.0 & 26.6 & 7.2 & 30.2 & 9.5 & 3.2 & -- & -- \\
LLaVA-Med & 7B & -- & -- & -- & -- & -- & -- & -- & -- & -- & -- & -- & 28.5 & 37.1 \\
Qwen2.5-VL & 3B & -- & -- & -- & -- & -- & -- & -- & -- & -- & -- & -- & 38.1 & 43.4 \\
HuatuoGPT-Vision & 7B & -- & -- & -- & -- & -- & -- & -- & -- & -- & -- & -- & 42.0 & 46.7 \\

\midrule

\rowcolor{ourrow}
\textbf{MedPixel} & 3B & 84.3 & 60.4 & 62.7 & 49.5 & 50.2 & 64.1 & \textbf{76.0} & 52.1 & 73.9 & 52.2 & 40.4 & 42.9 & 52.5 \\

\rowcolor{ourrow}
\textbf{MedPixel} & 7B & \textbf{85.0} & \textbf{61.7} & \textbf{66.7} & \textbf{53.4} & \textbf{51.3} & \textbf{70.8} & 75.9 & \textbf{53.1} & \textbf{76.5} & \textbf{55.2} & \textbf{42.3} & \textbf{47.5} & \textbf{58.1} \\

\bottomrule
\end{tabular}
\end{adjustbox}
\caption{
Comparison with representative methods across five tasks.
MET.\ denotes METEOR, and TGA denotes target grounding accuracy.
IMCQ and TMCQ denote image- and text-based multiple-choice accuracy, respectively.
``--'' denotes unsupported native interfaces.
For T3, results are averaged over point and box protocols.
}
\label{tab:main}
\end{table*}

\section{Experiments}
\label{sec:experiments}

\subsection{Experimental Setup}
\label{sec:setup}

\noindent\textbf{Training data.}
Stage~1 trains MedPixel on a 1.84M-sample mixture of medical VQA
(64\%), segmentation (32\%), and clinical reasoning (4\%) data.
MedPLG-440K provides the four segmentation-oriented tasks (T1--T4),
which are incorporated through task-specific sampling and replication.
The VQA component is aggregated from public medical QA datasets
~\cite{huatuogptvision,rocov2,pmcvqa,slake,pubmedqa,medmcqa,medqa,
medbullets}, while the reasoning component combines HuatuoGPT-o1,
Medical-R1-Distill, Citrus-S3~\cite{huatuogpto1,medicalr1distill,
citruss3}, and image-grounded reasoning synthesized from
PubMedVision~\cite{huatuogptvision}. The resulting checkpoint is used
to mine offline preference pairs for Stage~2. Full data statistics and training settings are provided in Appendix~C.

\noindent\textbf{Evaluation Sets.}
T1--T4 are derived from the held-out official BiomedParse test split, which is excluded from training. T1 evaluates explicit referring segmentation on 24,391 representative slices selected to reduce redundancy in volumetric studies. T2 and T4 contain 2,500 samples each for implicit reasoning and grounded explanation, respectively. T3 contains 2,000 interactive samples, with point and box prompts derived from ground-truth masks. T5 contains 10,267 image-based and text-based multiple-choice questions from public medical QA benchmarks. We further evaluate external transfer on the test sets of MeCoVQA-G+ and U-MRG-14K without benchmark-specific training.

\noindent\textbf{Metrics.}
For T1--T4, we report Dice and normalized surface Dice (NSD) with a 5-pixel tolerance; both are expressed as percentages, with the ``\%'' symbol omitted. For T2 and T4, we additionally report METEOR~\cite{meteor} for response
similarity. For T2, we also report target grounding accuracy (TGA), which measures whether the inferred target concept is fully recovered. Empty or invalid responses receive zero scores. For T5, we report multiple-choice accuracy. MeCoVQA-G+ is evaluated using slice-level micro and modality-level macro Dice, NSD, and HD95, while U-MRG-14K is evaluated using Dice. Full evaluation protocols are provided in Appendix~D.

\noindent\textbf{Baselines.}
We compare MedPixel with pixel-language models, segmentation specialists, and medical VQA models. Pixel-language baselines include LISA~\cite{lisa} and LISA++~\cite{lisa++}, PixelLM~\cite{pixellm}, SAM4MLLM~\cite{sam4mllm}, VisionReasoner~\cite{visionreasoner}, UniBiomed~\cite{unibiomed}, Citrus-V~\cite{citrusv}, MMedAgent~\cite{mmedagent}, MedPLIB~\cite{medplib}, and UniPixel~\cite{unipixel}. Segmentation specialists include BiomedParse~\cite{biomedparse}, MedSAM~\cite{medsam}, MedSAM2~\cite{medsam2}, SAM2~\cite{sam2}, and SAM-Med2D~\cite{sammed2d}. For T5, we additionally compare with LLaVA-Med~\cite{llavamed}, Qwen2.5-VL~\cite{qwen25vl} and HuatuoGPT-Vision~\cite{huatuogptvision}.

\noindent\textbf{Implementation details.}
We train MedPixel at 3B and 7B scales using Qwen2.5-VL as the multimodal backbone. LoRA~\cite{lora} is applied to the $q/k/v/o$ attention projections of both the language model and Qwen vision tower, using $r{=}128$ and $\alpha{=}256$, while their base weights remain frozen. Stage~1 optimizes the language adapters, multimodal projector, special token embeddings, segmentation projector, spatial reference modules, and SAM2 branch. Stage~2 updates the response-generation modules while freezing the segmentation branch.

\subsection{Main Results}
\label{sec:main_results}

Table~\ref{tab:main} compares MedPixel with segmentation specialists, pixel-language models, and medical VQA baselines. MedPixel achieves strong performance throughout the evaluated interfaces. On T1 referring segmentation, MedPixel-7B reaches $85.0$ Dice and $61.7$ NSD, exceeding BiomedParse by $5.0$ and $6.7$ points, respectively, showing that the unified framework retains strong explicit grounding performance. The advantage becomes substantially larger when localization requires semantic target inference, with improvements of $29.2$ Dice on T2 reasoning segmentation and $40.7$ Dice on T4 explanatory segmentation over the strongest baselines. MedPixel also achieves the highest METEOR on both tasks and the strongest TGA on T2, showing that the localization gains are achieved while preserving response similarity and target identification. On T3, MedPixel reaches $76.0$ Dice under its native interface combining semantic and spatial cues, remaining competitive with specialized promptable segmenters while supporting a broader interaction format. On T5, MedPixel-7B surpasses the strongest
baselines by $5.5$ and $7.0$ points on image- and text-based MCQs, respectively.

Scaling from 3B to 7B further improves MedPixel's overall capability, with gains across most tasks and particularly clear improvements on T2, T4, and T5. T1 improves modestly, while T3 remains essentially unchanged. This pattern suggests that increased model capacity strengthens semantic interpretation, target inference, and response generation, while the spatially guided interactive pathway is already strong at the 3B scale. Scaling enhances higher-level medical understanding without compromising dense grounding ability. Representative examples are shown in Figure~\ref{fig}.

\subsection{External Generalization}
\label{sec:external_generalization}

\noindent\textbf{MeCoVQA-G+.}
We evaluate zero-shot transfer on MeCoVQA-G+, which contains $2{,}719$ grounding samples across eight medical modalities. The benchmark is not used during training, and all results in Table~\ref{tab:mecovqa_external} are reproduced under the same evaluation protocol. Since MedPLIB is trained on MeCoVQA-G+, we report it only as a benchmark-trained reference rather than a zero-shot baseline. MedPixel-7B achieves the strongest overall performance among zero-shot text-driven models, improving over BiomedParse by $+5.3$ Dice at the slice level and $+3.4$ Dice at the modality level. It also reduces HD95 from $224.3/177.1$ to $152.3/123.0$, indicating more stable boundary localization across modalities.

\begin{table}[t]
\centering
\small
\begin{adjustbox}{width=\columnwidth}
\begin{tabular}{@{}l *{6}{c}@{}}
\toprule
\multicolumn{1}{c}{\multirow{2}{*}{\textbf{Method}}}
& \multicolumn{2}{c}{\textbf{Dice}$\uparrow$}
& \multicolumn{2}{c}{\textbf{NSD}$\uparrow$}
& \multicolumn{2}{c}{\textbf{HD95}$\downarrow$} \\
\cmidrule(lr){2-3}
\cmidrule(lr){4-5}
\cmidrule(l){6-7}
& \textbf{S} & \textbf{M}
& \textbf{S} & \textbf{M}
& \textbf{S} & \textbf{M} \\
\midrule
\rowcolor{ourrow}
MedPLIB-7B$^\dagger$ & 47.2 & 36.0 & 37.1 & 27.3 & 220.7 & 378.2 \\
\midrule
\textbf{MedPixel-7B} & \textbf{41.1} & \textbf{47.4} & \textbf{26.8} & \textbf{33.5} & \textbf{152.3} & \textbf{123.0} \\
\textbf{MedPixel-3B} & 40.5 & 47.3 & 26.2 & 33.2 & 163.5 & 132.5 \\
BiomedParse & 35.8 & 44.0 & 25.6 & 32.4 & 224.3 & 177.1 \\
UniBiomed-1B & 31.8 & 36.9 & 18.6 & 22.8 & 230.6 & 229.2 \\
VisionReasoner-7B & 26.5 & 33.3 & 12.0 & 18.0 & 291.7 & 243.6 \\
LISA-7B & 19.7 & 20.1 & 7.3 & 8.9 & 404.1 & 432.4 \\
LISA++-7B & 17.8 & 16.6 & 4.5 & 4.7 & 433.2 & 490.7 \\
MMedAgent-7B & 13.7 & 12.2 & 5.1 & 5.5 & 419.2 & 449.3 \\
SAM4MLLM-8B & 11.9 & 12.9 & 2.5 & 4.2 & 559.1 & 580.0 \\
PixelLM-7B & 11.1 & 10.4 & 1.9 & 1.9 & 539.7 & 567.4 \\
\bottomrule
\end{tabular}
\end{adjustbox}
\caption{Zero-shot transfer on MeCoVQA-G+. S/M denote slice-level micro and modality-level macro averages. Bold denotes the best zero-shot result. $^\dagger$ marks a benchmark-trained reference.}
\label{tab:mecovqa_external}
\end{table}

\begin{table}[t]
\centering
\small
\setlength{\tabcolsep}{1pt}
\renewcommand{\arraystretch}{1.06}
\begin{adjustbox}{max width=\columnwidth}
\begin{tabular}{@{}l r @{\hspace{8pt}} l r @{\hspace{8pt}} l r@{}}
\toprule
\multicolumn{2}{c}{\textbf{General MLLMs}}
& \multicolumn{2}{c}{\textbf{Medical MLLMs}}
& \multicolumn{2}{c}{\textbf{Grounding MLLMs}} \\
\cmidrule(lr){1-2}
\cmidrule(lr){3-4}
\cmidrule(l){5-6}
\textbf{Method} & \textbf{Dice}
& \textbf{Method} & \textbf{Dice}
& \textbf{Method} & \textbf{Dice} \\
\midrule
GPT-4o           & \phantom{0}4.72
& MedR1-2B        & 14.73
& VLMR1-REC-3B    & 22.19 \\
Gemini-2.5-flash & 14.29
& MiniInternVL-4B & \phantom{0}4.76
& SegZero-7B      & 26.05 \\
Qwen2.5-VL-7B    & 22.73
& MedGemma-4B     & \phantom{0}8.90
& SAM4MLLM-8B     & 16.49 \\
InternVL3-8B     & \phantom{0}9.23
& HuatuoGPT-7B    & 19.76
& MedReasoner-7B & 37.78 \\
Qwen2.5-VL-72B   & 29.71
& Lingshu-7B      & 16.48
& \cellcolor{ourrow}\textbf{MedPixel-7B}
& \cellcolor{ourrow}\textbf{37.91} \\
InternVL3-78B    & \phantom{0}7.23
& Chiron-o1-8B    & 10.05
&                  &       \\
\bottomrule
\end{tabular}
\end{adjustbox}
\caption{Zero-shot transfer on U-MRG-14K.}
\label{tab:umrg_external}
\end{table}

\noindent\textbf{U-MRG-14K.}
U-MRG-14K is a medical reasoning grounding benchmark introduced by MedReasoner~\cite{medreasoner}, containing $2{,}480$ test samples across ten modalities. Each sample pairs an implicit clinical query with a target mask. MedPixel is evaluated directly on the test set without benchmark-specific training. Results for the remaining methods in Table~\ref{tab:umrg_external} are taken from the published MedReasoner evaluation, where MLLMs generate spatial prompts for a fixed MedSAM2 segmenter. MedPixel-7B achieves comparable performance to MedReasoner-7B (37.91 vs. 37.78 Dice), while producing masks within a single unified model. This result further demonstrates that the resulting unified pixel-language model transfers effectively to unseen reasoning-grounding benchmarks.

\subsection{Ablation Study}
\label{sec:ablation}
We present the main findings here and defer further analyses to Appendix~E.

\noindent\textbf{Effect of PLPO.}
Table~\ref{tab:dpo_effect} shows that PLPO consistently improves Dice and NSD on T2 and T4 at both model scales, with larger gains on reasoning segmentation, while METEOR remains broadly stable. Since the segmentation projector and mask decoder are frozen during Stage~2, these improvements indicate better alignment between generated \texttt{<SEG>} representations and pixel-level outcomes. 

\begin{table}[t]
\centering
\setlength{\tabcolsep}{3.0pt}
\renewcommand{\arraystretch}{1.08}

\begin{adjustbox}{max width=\columnwidth}
\begin{tabular}{@{}c cc ccc ccc@{}}
\toprule
\multirow{2}{*}{\textbf{Scale}}
& \multirow{2}{*}{\textbf{SFT}}
& \multirow{2}{*}{\textbf{PLPO}}
& \multicolumn{3}{c}{\textbf{T2 Reasoning}}
& \multicolumn{3}{c}{\textbf{T4 Explanatory}} \\
\cmidrule(lr){4-6}
\cmidrule(l){7-9}
&
&
& Dice$\uparrow$ & NSD$\uparrow$ & MET.$\uparrow$
& Dice$\uparrow$ & NSD$\uparrow$ & MET.$\uparrow$ \\
\midrule

& \checkmark &
& 58.2 & 46.5 & \textbf{50.5}
& 72.1 & 51.2 & 39.7 \\

\rowcolor{ourrow}
\multirow{-2}{*}{\textbf{3B}}
& \checkmark & \checkmark
& \textbf{62.7} & \textbf{49.5} & 50.2
& \textbf{73.9} & \textbf{52.2} & \textbf{40.4} \\

\midrule

& \checkmark &
& 62.9 & 49.8 & 51.0
& 75.6 & 54.5 & 41.5 \\

\rowcolor{ourrow}
\multirow{-2}{*}{\textbf{7B}}
& \checkmark & \checkmark
& \textbf{66.7} & \textbf{53.4} & \textbf{51.3}
& \textbf{76.5} & \textbf{55.2} & \textbf{42.3} \\

\bottomrule
\end{tabular}
\end{adjustbox}
\caption{
Effect of Pixel-Level Preference Optimization (PLPO) on T2 and T4. MET.\ denotes METEOR.
}
\label{tab:dpo_effect}

\end{table}

\noindent\textbf{Dice--Reasoning Alignment.}
We use gpt-5.6-sol to determine whether each reasoning trace identifies the ground-truth target and compare these labels with Dice-based preferences. Reasoning-correct responses achieve much higher mean Dice than incorrect ones, both before PLPO ($84$ vs.\ $34$) and after PLPO ($87$ vs.\ $42$). Among preference pairs, $83.6\%$ favor a reasoning-correct response over an incorrect one, while only $0.4\%$ show the reverse, demonstrating strong alignment between Dice-based preferences and target-level reasoning.

\begin{figure}[t]
\centering
\includegraphics[width=0.95\columnwidth]{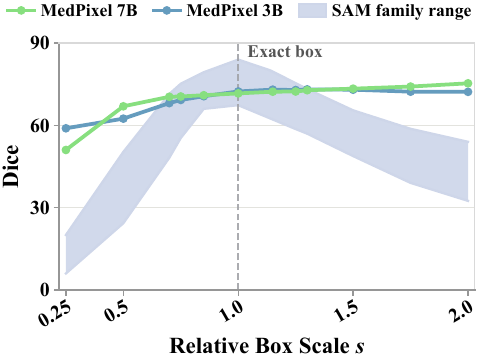}
\caption{Robustness to box scaling on T3.
Box width and height are scaled by $s$ with the center fixed; $s=1$ denotes the GT-tight box. The shaded region denotes the performance range of SAM-family baselines.}
\label{fig:box_robustness}
\end{figure}

\noindent\textbf{Robustness to Box Perturbations.}
To evaluate robustness to imperfect box prompts, we scale the width and height of each GT-tight box by a factor $s$ while keeping its center fixed and clipping the resulting box to the image boundaries. Here, $s=1$ denotes the exact box, whereas $s<1$ and $s>1$ produce tighter and looser prompts, respectively. Figure~\ref{fig:box_robustness} shows that SAM specialists perform strongly with exact boxes but degrade markedly as the box scale deviates from $1$. In contrast, MedPixel remains stable over a broad range of scales, demonstrating greater robustness to approximate spatial guidance. 

\section{Limitations}
MedPixel has several limitations. First, because MedPLG-440K is constructed by repurposing existing segmentation annotations, its coverage is limited to the anatomical structures and imaging modalities represented in the source datasets. Although the synthesis process requires neither additional annotation for individual samples nor external LLM annotation, language generated from rules and templates is less diverse than natural clinical interactions. Second, PLPO requires ground-truth masks for offline preference mining, limiting its applicability when pixel-level supervision is unavailable. Third, our evaluation mainly covers 2D or slice-based images and single-turn interactions, leaving volumetric, longitudinal, and multi-turn pixel--language reasoning underexplored. Finally, reference-based language metrics cannot fully assess the clinical correctness and completeness of generated explanations, calling for further evaluation by clinical experts.

\section{Conclusion}
\label{sec:conclusion}
This work presents \textbf{MedPixel}, a unified model for medical language
understanding, reasoning, interaction, and pixel-level localization.
\textbf{MedPLG-440K} converts existing segmentation annotations into diverse
pixel--language supervision, while \textbf{PLPO} uses mask quality to improve
the alignment between generated responses and segmentation outcomes. Across
five tasks, MedPixel achieves strong performance, with particularly clear
gains on reasoning and explanatory segmentation. It also generalizes well to
external benchmarks and remains robust to imperfect spatial prompts. Taken
together, these results demonstrate the feasibility of unifying medical
language reasoning, interaction, and pixel-level localization within a single
framework, with model conclusions grounded in precise visual evidence.

\bibliography{aaai2027}
\clearpage
\appendix

\twocolumn[
\begin{center}
    {\LARGE\bfseries Appendix}
\end{center}
\vspace{1em}
]
\begin{center}
\begin{minipage}{\textwidth}
\small
\noindent\textbf{Supplementary Material Overview}

\medskip
\begin{tabularx}{\textwidth}{@{}p{0.15\textwidth}X@{}}
\textbf{Appendix A} & Data Sources \\
\textbf{Appendix B} & MedPLG-440K Construction \\
\textbf{Appendix C} & Training Data and Optimization \\
\textbf{Appendix D} & Evaluation Protocols \\
\textbf{Appendix E} & Additional Experimental Results \\
\textbf{Appendix F} & Qualitative Results \\
\textbf{Appendix G} & Licenses \\
\end{tabularx}
\end{minipage}
\end{center}



\providecolor{ourrow}{RGB}{232,244,255}
\providecommand{\ourrow}{\rowcolor{ourrow}}
\providecommand{\na}{\textcolor{black!32}{--}}
\providecommand{\cmark}{\checkmark}
\providecommand{\xmark}{\ensuremath{\times}}
\newcommand{\slot}[1]{\textsf{\{#1\}}}
\newcommand{\segtoken}{\texttt{<|seg|>}}
\newcommand{\thinkopen}{\texttt{<think>}}
\newcommand{\thinkclose}{\texttt{</think>}}

\section{Data Sources}
\label{app:data_sources}

This section summarizes the data used to construct MedPLG-440K and
the auxiliary corpora used during Stage~1 training. Detailed dataset
construction and training-mixture composition are provided in
Appendices~\ref{app:construction} and~\ref{app:training}.

\subsection{Segmentation Data Sources}
\label{app:medplg_sources}

MedPLG-440K is constructed from the training annotations aggregated by
BiomedParse~\cite{biomedparse}. Table~\ref{tab:app_medplg_sources}
lists the corresponding original datasets according to the subset
mapping provided by BiomedParse.

\begin{table*}[t]
\centering
\footnotesize
\setlength{\tabcolsep}{4pt}
\renewcommand{\arraystretch}{1.08}
\begin{tabularx}{\textwidth}{
    @{}
    >{\raggedright\arraybackslash}p{0.145\textwidth}
    >{\raggedright\arraybackslash}X
    >{\raggedright\arraybackslash}p{0.145\textwidth}
    >{\raggedright\arraybackslash}X
    @{}
}
\toprule
\textbf{Subset}
& \textbf{Original dataset}
& \textbf{Subset}
& \textbf{Original dataset} \\
\midrule

ACDC
& Automated Cardiac Diagnosis Challenge
&
LGG
& LGG MRI Segmentation
\\

amos22
& AMOS 2022
&
LIDC-IDRI
& LIDC-IDRI
\\

BreastUS
& BUSI
&
LiverUS
& Ultrasound Simulation and Segmentation
\\

CAMUS
& CAMUS
&
MMs
& M\&Ms
\\

CDD-CESM
& CDD-CESM
&
MSD
& Medical Segmentation Decathlon
\\

COVID-19\_CT
& COVID-19 CT Segmentation Dataset
&
NeoPolyp
& BKAI-IGH NeoPolyp
\\

COVID-QU-Ex
& COVID-QU-Ex
&
OCT-CME
& Intraretinal Cystoid Fluid Dataset
\\

\makecell[l]{CXR\_Masks\_and\\Labels}
& Chest Xray Masks and Labels
&
PanNuke
& PanNuke
\\

DRIVE
& DRIVE
&
PolypGen
& PolypGen
\\

FH-PS-AOP
& FH-PS-AOP
&
QaTa-COV19
& QaTa-COV19
\\

G1020
& G1020
&
Radiography
& COVID-19 Radiography Database
\\

GlaS
& GlaS
&
REFUGE
& REFUGE
\\

ISIC
& ISIC 2018
&
\makecell[l]{siim-acr-\\pneumothorax}
& SIIM-ACR Pneumothorax
\\

kits23
& KiTS23
&
\makecell[l]{UWaterloo\\SkinCancer}
& UWaterloo Skin Cancer Dataset
\\

\bottomrule
\end{tabularx}
\caption{
Segmentation datasets underlying MedPLG-440K.
Subset identifiers and dataset mappings follow the BiomedParse
organization.
}
\label{tab:app_medplg_sources}
\end{table*}

\subsection{Medical QA and Clinical-Reasoning Sources}
\label{app:auxiliary_sources}

In addition to MedPLG-440K, Stage~1 incorporates image-based medical
VQA data from PubMedVision, ROCOv2, PMC-VQA, and SLAKE
~\cite{huatuogptvision,rocov2,pmcvqa,slake}; text-based QA data from
PubMedQA, MedMCQA, MedQA, and Medbullets
~\cite{pubmedqa,medmcqa,medqa,medbullets}; and clinical-reasoning data
from HuatuoGPT-o1, Medical-R1-Distill, and Citrus-S3
~\cite{huatuogpto1,medicalr1distill,citruss3}.
PubMedVision is additionally used to synthesize image-grounded
reasoning samples. Table~\ref{tab:app_auxiliary_sources} summarizes
these auxiliary sources.

\begin{table}[t]
\centering
\footnotesize
\setlength{\tabcolsep}{5pt}
\renewcommand{\arraystretch}{1.10}

\begin{tabularx}{\columnwidth}{
    @{}
    >{\raggedright\arraybackslash}p{0.31\columnwidth}
    >{\raggedright\arraybackslash}X
    @{}
}
\toprule
\textbf{Data group}
& \textbf{Dataset} \\
\midrule

\multirow{4}{*}{Image VQA}
& PubMedVision~\cite{pubmedqa} \\
& ROCOv2~\cite{rocov2} \\
& PMC-VQA~\cite{pmcvqa} \\
& SLAKE~\cite{slake} \\

\midrule

\multirow{4}{*}{Text QA}
& PubMedQA~\cite{pubmedqa} \\
& MedMCQA~\cite{medmcqa} \\
& MedQA~\cite{medqa} \\
& Medbullets~\cite{medbullets} \\

\midrule

\multirow{3}{*}{Clinical reasoning}
& HuatuoGPT-o1~\cite{huatuogpto1} \\
& Medical-R1-Distill~\cite{medicalr1distill} \\
& Citrus-S3~\cite{citruss3} \\

\bottomrule
\end{tabularx}

\caption{Medical question-answering and reasoning datasets used during
Stage~1 training.}
\label{tab:app_auxiliary_sources}
\end{table}

\section{MedPLG-440K Construction}
\label{app:construction}

MedPLG-440K contains $444{,}297$ pixel--language records constructed
from existing medical segmentation annotations across four tasks:
T1 Referring, T2 Reasoning, T3 Interactive, and T4 Explanatory.
The construction uses deterministic image--mask measurements, curated
clinical knowledge, and predefined templates, without external LLM
annotation.

\subsection{Dataset Composition}
\label{app:construction_composition}

Each source sample provides a medical image, its ground-truth mask,
the target category, and imaging modality. Visual attributes are first
extracted from the image--mask pair and then combined with
task-specific templates to generate four interaction formats.

T1 directly refers to a named target, T2 requires target inference
from indirect clinical clues, T3 introduces a point or box prompt,
and T4 combines segmentation with a medical explanation. A source
annotation may be reused across multiple formats. Table~
\ref{tab:app_constr_scale} summarizes the resulting composition.

\begin{table}[H]
\centering
\footnotesize
\begin{tabularx}{\columnwidth}{
    >{\raggedright\arraybackslash}l
    r
    >{\raggedright\arraybackslash}X}
\toprule
\textbf{Task} & \textbf{Records} & \textbf{Primary supervision} \\
\midrule
T1 Referring    & $94{,}261$  & Explicit target reference \\
T2 Reasoning    & $21{,}367$  & Indirect clinical query \\
T3 Interactive  & $318{,}670$ & Point or box guidance \\
T4 Explanatory  & $9{,}999$   & Grounded medical explanation \\
\midrule
\textbf{Total}  & $\mathbf{444{,}297}$ & \\
\bottomrule
\end{tabularx}
\caption{Composition of MedPLG-440K.}
\label{tab:app_constr_scale}
\end{table}

\subsection{Visual Attribute Extraction}
\label{app:constr_attributes}

For each image--mask pair, nine numerical measurements are
deterministically converted into seven language descriptors:
location, size, shape, components, boundary, contrast, and uniformity.
Let $I\in\mathbb{R}^{H\times W}$ denote the image,
$M\in\{0,1\}^{H\times W}$ the target mask, and $A=|M|$ its foreground
area.

\paragraph{Location.}
The mask centroid $(c_x,c_y)$ is assigned to a $3\times3$ spatial grid
using boundaries at $W/3$, $2W/3$, $H/3$, and $2H/3$. This produces
central, left, right, upper, lower, and four corner descriptors.

\paragraph{Size.}
The relative area
\begin{equation}
    r_A=\frac{A}{HW}
\end{equation}
is mapped to tiny, small, moderate-sized, large, or extensive using
cutoffs at $0.02$, $0.08$, $0.20$, and $0.45$.

\paragraph{Shape.}
For tight bounding-box dimensions $(w_b,h_b)$, we compute
\begin{equation}
    \rho=\frac{\max(w_b,h_b)}{\min(w_b,h_b)},
    \qquad
    \eta=\frac{A}{w_bh_b},
\end{equation}
where $\rho$ is the aspect ratio and $\eta$ the foreground fill ratio.
The target is classified as irregular if $\eta<0.45$; otherwise, it
is round if $\rho<1.3$, oval if $\rho<2.0$, and elongated otherwise.

\paragraph{Components.}
Let $K$ denote the number of connected foreground components. We use
single region for $K\leq1$, a few scattered foci for
$2\leq K\leq3$, and multiple scattered foci for $K\geq4$.

\paragraph{Boundary.}
From the largest external contour, we compute
\begin{equation}
    C=\frac{4\pi A_c}{P^2},
    \qquad
    V=\frac{A_c}{A_{\mathrm{hull}}},
    \qquad
    R=\frac{P}{\sqrt{A_c}},
\end{equation}
where $A_c$, $P$, and $A_{\mathrm{hull}}$ denote the contour area,
perimeter, and convex-hull area. The boundary score is
\begin{equation}
\begin{aligned}
s={}&2\mathbb{I}(C>0.75)
+\mathbb{I}(0.55<C\leq0.75)\\
&+2\mathbb{I}(V>0.90)
+\mathbb{I}(0.75<V\leq0.90)\\
&+2\mathbb{I}(R<4.5)
+\mathbb{I}(4.5\leq R<6.5),
\end{aligned}
\label{eq:app_boundary_score}
\end{equation}
where $\mathbb{I}(\cdot)$ denotes the indicator function. Scores
$s\geq5$, $s=4$, $s=3$, $s=2$, and $s<2$ correspond to
well-circumscribed and smooth margins, well-defined margins,
partially defined margins, lobulated and partially irregular margins,
and irregular and ill-defined margins, respectively.

\paragraph{Contrast.}
The foreground-to-background intensity ratio
\begin{equation}
    I_r=\frac{\mu_{\mathrm{fg}}}{\mu_{\mathrm{bg}}}
\end{equation}
is classified as high for $I_r>1.15$, low for $I_r<0.85$, and iso
otherwise. These levels are expressed using the modality-specific
terms in Table~\ref{tab:app_modality_terms}.

\begin{table}[t]
\centering
\footnotesize
\resizebox{\columnwidth}{!}{%
\begin{tabular}{llll}
\toprule
\textbf{Modality} & \textbf{High} & \textbf{Iso} & \textbf{Low} \\
\midrule
CT          & hyperdense      & isodense   & hypodense \\
MRI         & hyperintense    & isointense & hypointense \\
Ultrasound  & hyperechoic     & isoechoic  & hypoechoic \\
X-ray       & radiopaque      & ---        & radiolucent \\
Mammography & hyperdense      & isodense   & hypodense \\
OCT         & hyperreflective & ---        & hyporeflective \\
\bottomrule
\end{tabular}%
}
\caption{Modality-specific terminology for relative intensity.}
\label{tab:app_modality_terms}
\end{table}

\paragraph{Uniformity.}
The foreground intensity standard deviation
$\sigma_{\mathrm{fg}}$ is mapped to homogeneous for
$\sigma_{\mathrm{fg}}<12$, relatively uniform for
$\sigma_{\mathrm{fg}}<30$, mildly heterogeneous for
$\sigma_{\mathrm{fg}}<55$, and markedly heterogeneous otherwise.
For homogeneous regions, the wording is adapted to the modality:
internal echotexture for ultrasound, attenuation for CT, density for
mammography, signal intensity for MRI, and internal texture otherwise.

\begin{table*}[t]
\centering
\footnotesize
\setlength{\tabcolsep}{8pt}
\renewcommand{\arraystretch}{1.10}
\begin{tabularx}{\textwidth}{
    >{\raggedright\arraybackslash}p{0.14\textwidth}
    >{\raggedright\arraybackslash}X}
\toprule
\textbf{Task} & \textbf{Instruction and response} \\
\midrule

T1 Referring
&
\textbf{User:} Segment the liver.

\smallskip
\textbf{Assistant:} The target is <|seg|>.
\\
\addlinespace[3pt]

T2 Reasoning
&
\textbf{User:} Where is the large solid organ in the right upper quadrant of the
abdomen? Segment it.

\smallskip
\textbf{Assistant:} <think> This requires anatomical reasoning to determine the
segmentation target. The query specifies the large solid organ in the
right upper quadrant of the abdomen as the defining landmark. The
liver is the structure that fits the description of the large solid
organ in the right upper quadrant of the abdomen. I will now produce
a precise mask for the liver. </think> 

This is the liver, located as
described. I can see the liver in the central region. It is a large,
oval structure displaying well-circumscribed, smooth margins,
hyperechoic relative to surrounding tissue, with mildly heterogeneous
texture. <|seg|>
\\
\addlinespace[3pt]

T3 Interactive
&\textbf{User:} Find the edges of the organ here~<|ref|> and tell me how big it is.

\smallskip
\textbf{Assistant:} This is the liver in abdominal ultrasound. It presents in the central
region as a large, oval structure. <|seg|>
\\
\addlinespace[3pt]

T4 Explanatory
&
\textbf{User:} Provide a clinical assessment of this ultrasound and segment the
relevant structures.

\smallskip
\textbf{Assistant:} <think> My task is to identify the liver in this ultrasound based on
its imaging characteristics. I can distinguish the liver from
adjacent structures based on its typical anatomical position and
imaging appearance. The segmentation will follow the margin of the
liver, capturing its complete extent in this view. </think> 

On careful review, the liver is visible in the central region as a large, oval
structure with well-circumscribed, smooth margins, hyperechoic
relative to surrounding tissue, with mildly heterogeneous texture.
It receives dual blood supply from the hepatic artery and portal
vein. <|seg|>
\\

\bottomrule
\end{tabularx}
\caption{Representative MedPLG-440K examples.}
\label{tab:app_constr_examples}
\end{table*}

\subsection{Task-Specific Synthesis}
\label{app:constr_synthesis}

The extracted visual information is incorporated into predefined
task-specific templates together with category names, imaging
modalities, clinical clues, and spatial prompts. T1 directly uses the
category name, T2 and T4 use the complete descriptor set, and T3 uses
only location, size, and shape. Unavailable contrast and uniformity
clauses are omitted automatically.

\paragraph{T1 Referring}
T1 explicitly names the target category. Its prompt bank covers seven
styles, including direct commands, polite requests, questions,
technical expressions, and presence checks. A representative prompt
is ``Segment the \{category\}.'' The response is sampled from five
short templates, such as ``Sure, the segmentation mask is <|seg|>.''

\paragraph{T2 Reasoning}
T2 omits the category name and describes the target through functional,
anatomical, symptomatic, diagnostic, exclusion-based, or
appearance-based clues. A representative prompt is
``Segment the structure responsible for \{function\}.'' The response
contains a template-constructed reasoning trace, a conclusion
identifying the inferred target, and an observation instantiated from
the extracted descriptors:
\begin{quote}
\small
<think>
\{opening\} $\rightarrow$
\{clue interpretation\} $\rightarrow$
\{category inference\} $\rightarrow$
\{segmentation approach\}
</think>

\{conclusion\} The \{category\} occupies the \{location\} region as a
\{size\} \{components\} with \{shape\} contours and
\{boundary\}\{contrast clause\}\{uniformity clause\}. <|seg|>
\end{quote}

\paragraph{T3 Interactive}
T3 pairs an instruction with either a positive foreground point or a
tight bounding box. Point and box prompts are sampled from separate
template banks, for example,
``Segment the foreground object marked by the point.'' For non-organ
categories, organ is replaced with structure. The response identifies
the indicated target and describes its location, size, and shape:

\begin{quote}
\small
This structure is the \{category\}, situated in the \{location\}
region with a \{size\}, \{shape\} appearance. <|seg|>
\end{quote}

\paragraph{T4 Explanatory}
T4 uses explicit, implicit, and clinical prompt families with
probabilities $0.40$, $0.30$, and $0.30$, respectively. Explicit
prompts name the target, implicit prompts request pathology detection,
and clinical prompts request general image analysis. Implicit prompts
fall back to the explicit form for non-pathological targets.

The response combines a template-constructed reasoning trace, a
visual observation, and an optional medical knowledge
sentence:
\begin{quote}
\small
<think>
\{opening\} $\rightarrow$
\{visual analysis\} $\rightarrow$
\{segmentation approach\} $\rightarrow$
\{optional modality cue\}
</think>

\{prefix\} The \{category\} is visible in the \{location\} region as
a \{size\}, \{shape\} structure with \{components\} and
\{boundary\}\{contrast clause\}\{uniformity clause\}.
\{medical knowledge\} <|seg|>
\end{quote}

Template diversity is further increased through predefined clinical
phrases, synonym substitution for T2 clues, modality-specific
terminology, and automatic article correction. Table~
\ref{tab:app_constr_examples} presents verbatim examples generated
from a shared liver annotation.

\section{Training Data and Optimization}
\label{app:training}

\subsection{Stage~1 Training Mixture}
\label{app:stage1_mixture}

Stage~1 combines MedPLG-440K, public medical VQA and QA corpora, and
distilled clinical-reasoning data. Table~\ref{tab:app_stage1_composition}
summarizes the effective mixture after task-level replication.

\begin{table}[t]
\centering
\footnotesize
\begin{tabularx}{\columnwidth}{
    >{\raggedright\arraybackslash}X
    r
    r}
\toprule
\textbf{Data source} & \textbf{Instances} & \textbf{Share} \\
\midrule
MedPLG-440K grounding
& $591{,}291$
& $32\%$ \\
Medical VQA and QA
& $1{,}178{,}760$
& $64\%$ \\
Distilled clinical reasoning
& $72{,}751$
& $4\%$ \\
\midrule
\textbf{Total}
& $\mathbf{1{,}842{,}802}$
& $\mathbf{100\%}$ \\
\bottomrule
\end{tabularx}
\caption{Effective Stage~1 training mixture.}
\label{tab:app_stage1_composition}
\end{table}

T1 and T4 are repeated twice, T2 three times, and T3 once, yielding

\begin{equation}
\begin{aligned}
N_{\mathrm{ground}}
={}&2\times94{,}261
+3\times21{,}367\\
&+318{,}670
+2\times9{,}999\\
={}&591{,}291.
\end{aligned}
\end{equation}

Replication changes only sampling frequency; MedPLG-440K retains
$444{,}297$ unique records.

\subsection{Stage~1 Supervised Fine-Tuning}
\label{app:stage1_training}

The 3B and 7B variants use the same Stage~1 training configuration.
For the joint objective defined in the main paper, we set
$\lambda_{\mathrm{seg}}=1$,
$\lambda_{\mathrm{focal}}=100$, and
$\lambda_{\mathrm{dice}}
=\lambda_{\mathrm{iou}}
=\lambda_{\mathrm{obj}}=5$.
The focal loss uses $\alpha=0.25$ and $\gamma=2$. Samples without
pixel-level annotations are optimized only with the language modeling
loss.

The LoRA dropout is set to $0.1$. The language- and vision-side LoRA
adapters, multimodal and task-specific projection modules, token
embeddings, and language output head use a learning rate of
$1.5\times10^{-5}$. The SAM2 image encoder, prompt encoder, and mask
decoder are jointly optimized with a smaller learning rate of
$5\times10^{-6}$.

Training uses AdamW for one epoch with a cosine learning-rate schedule
and a $3\%$ warmup ratio. We use a per-device batch size of $1$,
gradient accumulation over $4$ steps, a maximum sequence length of
$4{,}096$, and bf16 precision.

\subsection{Stage~2 Pixel-Level Preference Optimization}
\label{app:plpo_details}

\paragraph{Candidate generation.}
For each T2 reasoning-segmentation input, the Stage~1 policy generates
$N=8$ candidate responses: one greedy decode, three stochastic samples
with temperature $0.3$ and top-$p=0.9$, and four samples with
temperature $0.7$ and top-$p=0.95$. The maximum generation length is
$1{,}024$ new tokens.

\paragraph{Pair construction.}
Following the filtering and scoring procedure described in the main
paper, we set $\tau_{\min}=0.30$ and
$\tau_{\mathrm{gap}}=0.10$. When multiple lower-scoring candidates
satisfy the required Dice margin, the highest-scoring one is retained
as the rejected response. Inputs without a qualified pair are
discarded, resulting in $1{,}786$ preference pairs from the T2
training branch.

\paragraph{Optimization settings.}
The Stage~1 checkpoint initializes the trainable policy and serves as
the frozen reference policy. Reference log-probabilities are
precomputed and cached before training. We set
$\beta=0.25$ and $\lambda_{\mathrm{nll}}=0.05$, without response-length
normalization.

Stage~2 is trained for three epochs with a learning rate of
$1\times10^{-6}$. The per-device batch size is $1$ with gradient
accumulation over $4$ steps. The maximum DPO sequence length is
$2{,}048$, and the maximum gradient norm is $0.5$.

The original language and visual backbone weights remain frozen.
Stage~2 optimizes their $q/k/v/o$ LoRA adapters, together with the
visual merger, token embeddings, and language output head. The
segmentation projector and the complete SAM2 branch remain frozen.

All training is conducted on NVIDIA RTX PRO 6000 GPU.

\section{Evaluation Protocols}
\label{app:evaluation}

\subsection{In-Domain Evaluation Sets}
\label{app:indomain_evaluation}

T1--T4 are constructed from the official BiomedParse test directories,
without introducing an additional train--test split. Sequence
construction is performed independently within each official split.

\paragraph{Clip construction.}
Eight volumetric datasets, including ACDC, MSD, AMOS22, KiTS23, LGG,
LIDC-IDRI, MMs, and COVID-19 CT, are treated as 3D data. Their slices
are grouped by volume identity and target category, ordered by slice
index, and divided when the gap between adjacent annotated slices is
at least three. Sequences longer than 16 frames are further divided
using a window length of 16 and a stride of 8. Intermediate slices
without target annotations are retained as negative frames with empty
masks. Images from the remaining 2D datasets are represented as
single-frame clips.

This procedure produces $24{,}391$ test clips, comprising $17{,}489$
single-frame clips and $6{,}902$ multi-frame clips.

\paragraph{T1 evaluation.}
T1 evaluates all $24{,}391$ test clips using one frame per clip. For a
clip with $L$ frames, the frame at index $\lfloor L/2 \rfloor$ is
selected. If the middle frame does not contain a valid target mask, an
annotated frame from the same clip is used instead.

\paragraph{T2--T4 evaluation.}
T2, T3, and T4 are sampled from the official test pool using
dataset-balanced sampling with a fixed random seed of $123$.
Candidates must be single-frame samples with valid masks, and duplicate
frame--mask pairs are removed. T2 further restricts targets to the
predefined organ-knowledge set. The resulting sets contain $2{,}500$
T2 samples, $2{,}000$ T3 samples, and $2{,}500$ T4 samples. T3
contains equal numbers of point-prompted and box-prompted examples.

\paragraph{T5 evaluation.}
T5 contains $10{,}267$ multiple-choice questions from public medical
QA benchmarks, including $1{,}647$ image-based and $8{,}620$
text-based samples.

\subsection{External Evaluation}
\label{app:external_evaluation}

We evaluate zero-shot transfer on the official test sets of
MeCoVQA-G+ and U-MRG-14K, containing $2{,}719$ and $2{,}480$
evaluation samples, respectively. No benchmark-specific fine-tuning
or preference optimization is performed.

For MeCoVQA-G+, slice-level results are aggregated over all samples to
obtain micro scores, while modality-level scores are computed
separately and averaged to obtain macro performance. U-MRG-14K is
evaluated using its official segmentation protocol.

\subsection{Evaluation Metrics}
\label{app:evaluation_metrics}

For a predicted mask $\hat{M}$ and ground-truth mask $M$, Dice is
computed as
\begin{equation}
\operatorname{Dice}(\hat{M},M)
=
\frac{2|\hat{M}\cap M|}
{|\hat{M}|+|M|}.
\end{equation}
Normalized surface Dice is computed with a boundary tolerance of five
pixels. HD95 measures the $95$th-percentile bidirectional surface
distance, with lower values indicating better boundary agreement.

For T2 and T4, METEOR~\cite{meteor} measures response similarity,
while target grounding accuracy measures whether the normalized target
concept is fully recovered. Empty or invalid responses and outputs
without a decodable mask receive zero scores.

For T5, the predicted option is extracted from the generated response
and evaluated using multiple-choice accuracy.

\section{Additional Experimental Results}
\label{app:additional_results}

\subsection{Detailed Interactive Segmentation Results}
\label{app:t3_detailed_results}

T3 interactive segmentation contains two equally sized subsets:
$1{,}000$ point-prompted samples and $1{,}000$ box-prompted samples.
While the main paper reports the average performance over the two
subsets, Table~\ref{tab:t3_prompt_results} presents the separate Dice
and NSD results.

\begin{table}[t]
\centering
\begin{tabular}{@{}ccccc@{}}
\toprule
\multirow{2}{*}{\textbf{Prompt}}
& \multicolumn{2}{c}{\textbf{MedPixel-3B}}
& \multicolumn{2}{c}{\textbf{MedPixel-7B}} \\
\cmidrule(lr){2-3}
\cmidrule(lr){4-5}
& \textbf{Dice$\uparrow$}
& \textbf{NSD$\uparrow$}
& \textbf{Dice$\uparrow$}
& \textbf{NSD$\uparrow$} \\
\midrule
Box
& 72.36 & 49.39
& 71.65 & 49.79 \\
Point
& 79.65 & 54.88
& 80.16 & 56.47 \\
\midrule
Avg.
& 76.00 & 52.14
& 75.91 & 53.13 \\
\bottomrule
\end{tabular}
\caption{
Detailed T3 interactive-segmentation results under point and box
prompts. Each subset contains $1{,}000$ samples. Avg. denotes the
arithmetic mean over the two prompt types.
}
\label{tab:t3_prompt_results}
\end{table}

As shown in Table~\ref{tab:t3_prompt_results}, point prompts
consistently outperform box prompts for both model scales. For
MedPixel-3B, point prompting improves Dice and NSD by $7.29$ and
$5.49$ points, respectively, while MedPixel-7B achieves gains of
$8.51$ Dice and $6.68$ NSD. MedPixel-7B obtains the strongest
point-prompted performance, reaching $80.16$ Dice and $56.47$ NSD.

The stronger performance under point prompts is also encouraging for
practical use, as a point can be specified with a single click and
requires less interaction effort than drawing a bounding box. These
results suggest that MedPixel can achieve accurate interactive
segmentation from lightweight spatial guidance.

\subsection{Detailed Effect of Pixel-Level Preference Optimization}
\label{app:plpo_effect}

We provide a detailed comparison between the Stage~1 SFT and
Stage~2 PLPO checkpoints across all five tasks. For T1--T4, we report
Dice, gIoU, cIoU, and NSD. For the language-generative T2 and T4 tasks,
we additionally report METEOR (MET.) and target grounding accuracy
(TGA). TGA measures whether all tokens in the reference target phrase
are recovered in the generated response. For T5, IMCQ and TMCQ denote
image-based and text-based medical multiple-choice accuracy,
respectively. Higher values are better for all metrics. Dashes indicate
that the corresponding NSD results were not retained during the earlier
3B Stage~1 evaluation.

\begin{table*}[t]
\centering

\renewcommand{\arraystretch}{1.10}
\begin{adjustbox}{max width=\textwidth}
\begin{tabular}{
@{}c l
*{4}{c}
*{6}{c}
*{4}{c}
*{6}{c}
*{2}{c}
@{}
}
\toprule
\multirow{2}{*}{\textbf{Scale}}
& \multirow{2}{*}{\textbf{Training}}
& \multicolumn{4}{c}{\textbf{T1 Referring}}
& \multicolumn{6}{c}{\textbf{T2 Reasoning}}
& \multicolumn{4}{c}{\textbf{T3 Interactive}}
& \multicolumn{6}{c}{\textbf{T4 Explanatory}}
& \multicolumn{2}{c}{\textbf{T5 VQA}} \\
\cmidrule(lr){3-6}
\cmidrule(lr){7-12}
\cmidrule(lr){13-16}
\cmidrule(lr){17-22}
\cmidrule(lr){23-24}
&
& Dice & gIoU & cIoU & NSD
& Dice & gIoU & cIoU & NSD & MET. & TGA
& Dice & gIoU & cIoU & NSD
& Dice & gIoU & cIoU & NSD & MET. & TGA
& IMCQ & TMCQ \\
\midrule

& Stage~1
& 84.29 & 76.95 & \textbf{82.15} & 60.39
& 58.19 & 50.99 & 47.92 & 46.45 & \textbf{50.5} & 62.9
& 75.72 & 67.94 & 72.69 & \textbf{52.16}
& 72.12 & 65.35 & 67.06 & 51.19 & 39.7 & 82.4
& 42.14 & \textbf{52.46} \\

\rowcolor{ourrow}
\multirow{-2}{*}{\textbf{3B}}
& \textbf{Stage~2}
& \textbf{84.30} & \textbf{76.96} & \textbf{82.15} & 60.40
& \textbf{62.70} & \textbf{55.09} & \textbf{53.50} & 49.46
& 50.2 & \textbf{64.1}
& \textbf{76.00} & \textbf{68.17} & \textbf{72.90} & 52.14
& \textbf{73.93} & \textbf{66.93} & \textbf{69.07} & \textbf{52.18}
& \textbf{40.4} & \textbf{83.4}
& \textbf{42.87} & 52.45 \\

\midrule

& Stage~1
& \textbf{85.02} & \textbf{77.83} & \textbf{83.00} & 61.65
& 62.90 & 55.83 & 53.48 & 49.80 & 51.0 & \textbf{72.5}
& \textbf{75.97} & \textbf{68.84} & \textbf{74.34} & \textbf{53.22}
& 75.63 & 68.59 & 71.28 & 54.50 & 41.5 & 87.6
& 47.30 & 57.84 \\

\rowcolor{ourrow}
\multirow{-2}{*}{\textbf{7B}}
& \textbf{Stage~2}
& 85.00 & 77.81 & 82.97 & \textbf{61.67}
& \textbf{66.65} & \textbf{59.20} & \textbf{58.27} & \textbf{53.35}
& \textbf{51.3} & 70.8
& 75.91 & 68.83 & 73.75 & 53.13
& \textbf{76.46} & \textbf{69.42} & \textbf{71.96} & \textbf{55.24}
& \textbf{42.3} & \textbf{87.9}
& \textbf{47.48} & \textbf{58.14} \\

\bottomrule
\end{tabular}
\end{adjustbox}
\caption{
Detailed effect of Stage~2 Pixel-Level Preference Optimization across
all five tasks. MET. denotes METEOR, and TGA denotes target grounding
accuracy.
}
\label{tab:plpo_effect_detailed}
\end{table*}

Table~\ref{tab:plpo_effect_detailed} shows that PLPO produces its
largest improvements on T2 reasoning segmentation. For MedPixel-3B,
Dice, gIoU, and cIoU increase by $4.51$, $4.10$, and $5.58$ points,
respectively. The corresponding gains for MedPixel-7B are $3.75$,
$3.37$, and $4.79$ points, together with a $3.55$-point improvement
in NSD. These consistent changes show that PLPO improves both
sample-level overlap and dataset-level pixel aggregation.

T4 explanatory segmentation also benefits from PLPO. For MedPixel-3B,
Dice, gIoU, and cIoU improve by $1.81$, $1.58$, and $2.01$ points,
respectively. MedPixel-7B obtains smaller but consistent improvements
across all four segmentation metrics. METEOR and TGA also improve on
T4 for both model scales, indicating that the pixel-level gains do not
degrade the generated explanations.

On T2, the response metrics remain broadly stable. METEOR changes by
only $-0.3$ and $+0.3$ points for the 3B and 7B models, respectively,
while TGA changes by $+1.2$ and $-1.7$ points. These variations are
small relative to the segmentation gains, suggesting that PLPO mainly
improves the pixel-level outcomes while preserving the learned
reasoning behavior.

In contrast, T1 referring segmentation and T3 interactive segmentation
remain nearly unchanged after PLPO. Their Dice, gIoU, cIoU, and NSD
values exhibit only minor fluctuations, with no consistent degradation.
T5 medical VQA is similarly preserved: IMCQ improves at both scales,
while TMCQ remains nearly unchanged for MedPixel-3B and improves
slightly for MedPixel-7B.

Overall, PLPO selectively improves reasoning- and
explanation-oriented segmentation while maintaining referring
segmentation, interactive segmentation, and medical question-answering
performance.

\subsection{Dice--Reasoning Alignment}
\label{app:plpo_alignment}

We conduct an additional analysis to examine whether the pixel-level
preference signal used by PLPO is consistent with target-level reasoning.
The analysis covers the $2{,}500$ responses generated on the T2 test set
before and after PLPO, together with the $1{,}786$ mined preference pairs.
The headline comparisons are reported in the main paper; here, we describe
the reasoning-annotation protocol and provide complementary statistics.

\noindent\textbf{Reasoning annotation protocol.}
We use gpt-5.6-sol to determine whether each generated
\textless think\textgreater trace identifies the ground-truth segmentation target. For each response, the judge receives the segmentation query, the ground-truth target, and the generated reasoning trace. The image, predicted mask, and Dice score are excluded from the judging input, so the reasoning label is assigned independently of segmentation quality.

The judge returns one of three labels: \emph{yes}, \emph{partial}, or
\emph{no}. A response is labeled \emph{yes} when its concluded target matches
the ground-truth structure, allowing synonyms, clinically equivalent expressions, and equivalent image- and patient-side laterality descriptions.
A \emph{partial} label denotes the correct organ with an incorrect
substructure or a genuinely ambiguous conclusion. A \emph{no} label denotes
a different structure, incorrect laterality, or the absence of an
identifiable target. Only \emph{yes} is counted as reasoning-correct;
\emph{partial}, \emph{no}, and unparseable outputs are treated as incorrect.
The same fixed rubric is applied to the SFT and PLPO outputs and to both
responses in every preference pair.

The exact judging prompt is reproduced below.

\begin{quote}
\small
\textbf{System prompt.}
You are a strict judge of medical image-segmentation reasoning. For each item
you get: QUESTION (what to segment), GT (the ground-truth target structure
whose mask is the correct answer), and REASONING (the model's
\texttt{<think>} text). Decide whether the REASONING concludes a target that
is the SAME structure as GT. Allow synonyms, laterality equivalence between
image-side and patient-side wording, and clinical equivalence. Judge IDENTITY
ONLY against GT; ignore mask and segmentation quality, and ignore whether the
wording in QUESTION appears to contradict GT, since GT is authoritative.

Label \texttt{yes} when the concluded target equals GT or an equivalent
structure; label \texttt{partial} when the correct organ is identified but
the substructure is incorrect, or when the conclusion is genuinely ambiguous;
and label \texttt{no} when the response concludes a different structure,
uses incorrect laterality, or contains no clear target. Return only a compact
JSON array, with one object per item and no additional prose:
\{"uid":"...","c":"yes|no|partial","t":"concluded target"\}.

\medskip
\textbf{User-message template.}
Judge these items:
\{"uid":"...","question":"<segmentation query>",\\
"gt":"<ground-truth target>",\\
"reasoning":"<generated reasoning trace>"\}.
\end{quote}

\begin{figure*}[t]
\centering
\includegraphics[width=\textwidth]{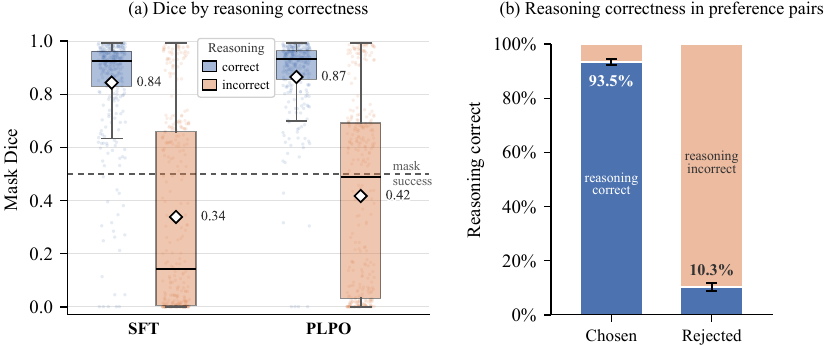}
\caption{
Dice--reasoning alignment on T2.
\textbf{(a)} Reasoning-correct responses achieve substantially higher Dice
before and after PLPO.
\textbf{(b)} Higher-Dice chosen responses are substantially more likely to
be reasoning-correct than rejected responses.
}
\label{fig:app_plpo_alignment}
\label{fig:app_plpo_alignment}
\end{figure*}

\noindent\textbf{Response-level association.}
As shown in Fig.~\ref{fig:app_plpo_alignment}(a), reasoning-correct
responses decode substantially better masks under both policies. Their mean
Dice scores are $84.4$ versus $33.8$ for SFT and $86.6$ versus $41.7$
after PLPO. Treating Dice as a score for distinguishing reasoning-correct
from reasoning-incorrect responses yields AUC values of $0.861$ and $0.842$,
respectively. The consistently high AUC indicates that the association
between target-level reasoning and mask quality remains strong after
preference optimization.

\noindent\textbf{Preference-level association.}
Figure~\ref{fig:app_plpo_alignment}(b) shows that $93.5\%$ of chosen
responses are reasoning-correct, compared with $10.3\%$ of rejected
responses. At the pair level, $83.6\%$ prefer a reasoning-correct response
over an incorrect one, whereas only $0.4\%$ exhibit the reverse ordering.
Among pairs whose candidates differ in reasoning correctness, the
higher-Dice response is correct in $99.5\%$ of cases. Thus, the preference
ordering induced by mask quality is highly consistent with target-level
reasoning.

\noindent\textbf{Reasoning preservation after PLPO.}
On the same $2{,}500$ T2 test samples, PLPO increases mean Dice from $62.9$
to $66.7$, while the reasoning-correct rate changes from $57.4\%$ to
$55.6\%$. This difference is not statistically significant under McNemar's
test ($p=0.07$). Moreover, reasoning changes from correct to incorrect while
Dice improves in only $1.4\%$ of samples. These results indicate that PLPO
improves segmentation without a detectable systematic degradation in
target-level reasoning.

\subsection{Detailed Evaluation of Generated Responses}
\label{app:detailed_response_evaluation}

\paragraph{Evaluation protocol.}
We further evaluate the pixel-level predictions and generated responses
on the $2{,}500$-sample T2 reasoning-segmentation and T4
explanatory-segmentation test sets. For T2, we retain the complete
reasoning and final answer after removing structural tags,
segmentation tokens, coordinate tokens, and chat templates. For T4,
we evaluate the final explanatory text after the reasoning block.

\paragraph{Evaluation metrics.}
Dice and NSD evaluate region overlap and boundary agreement,
respectively. Avg. Len. reports the average number of content words
after text cleaning. METEOR (MET.) measures similarity to
the reference response. Target grounding accuracy (TGA) requires all
tokens in the reference target phrase to be recovered, while Head
Match requires recovery of its principal anatomical or finding term. Avg. Len. has no
preferred direction.

\begin{table*}[t]
\centering
\renewcommand{\arraystretch}{1.10}
\begin{adjustbox}{max width=\textwidth}
\begin{tabular}{@{}l *{12}{c}@{}}
\toprule
\multirow{2}{*}{\textbf{Method}}
& \multicolumn{6}{c}{\textbf{T2 Reasoning Segmentation}}
& \multicolumn{6}{c}{\textbf{T4 Explanatory Segmentation}} \\
\cmidrule(lr){2-7}
\cmidrule(lr){8-13}
& \textbf{Dice$\uparrow$}
& \textbf{NSD$\uparrow$}
& \textbf{Len.}
& \textbf{MET.$\uparrow$}
& \textbf{TGA$\uparrow$}
& \textbf{Head$\uparrow$}
& \textbf{Dice$\uparrow$}
& \textbf{NSD$\uparrow$}
& \textbf{Len.}
& \textbf{MET.$\uparrow$}
& \textbf{TGA$\uparrow$}
& \textbf{Head$\uparrow$} \\
\midrule

Citrus-V-8B
& 37.5 & 25.9 & -- & -- & -- & --
& 35.1 & 23.0 & 16 & 10.2
& \textbf{100.0} & \textbf{100.0} \\

VisionReasoner-7B
& 33.4 & 14.5 & 87 & 29.6 & 39.8 & 47.4
& 31.1 & 13.0 & 182 & 21.2 & 72.0 & 76.9 \\

UniBiomed-1.4B
& 28.3 & 14.4 & 86 & 4.9 & 27.3 & 29.9
& 35.8 & 19.1 & 55 & 7.3 & 23.7 & 25.9 \\

LISA++-7B
& 22.8 & 8.9 & 6 & 7.9 & 0.1 & 0.2
& 12.1 & 5.3 & 46 & 12.0 & 9.3 & 11.1 \\

PixelLM-7B
& 18.2 & 3.0 & 19 & 8.1 & 4.9 & 5.8
& 10.0 & 1.5 & 30 & 11.6 & 14.2 & 15.1 \\

UniPixel-7B
& 27.9 & 7.5 & 2 & 3.0 & 0.0 & 0.0
& 30.2 & 9.5 & 5 & 3.2 & 0.0 & 0.0 \\

\midrule

\rowcolor{ourrow}
MedPixel-3B
& 62.7 & 49.5 & 92
& 50.2 & 64.1 & 73.9
& 73.9 & 52.2 & 44
& 40.4 & 83.4 & 87.8 \\

\rowcolor{ourrow}
\textbf{MedPixel-7B}
& \textbf{66.7} & \textbf{53.4} & 92
& \textbf{51.3} & \textbf{70.8} & \textbf{79.2}
& \textbf{76.5} & \textbf{55.2} & 42
& \textbf{42.3} & 87.9 & 91.0 \\

\bottomrule
\end{tabular}
\end{adjustbox}
\caption{
Detailed evaluation of pixel-level predictions and generated responses
on T2 reasoning segmentation and T4 explanatory segmentation.
Avg. Len. is measured in content words, and TGA denotes target
grounding accuracy based on recovery of all target tokens.
}
\label{tab:t2_t4_detailed_response}
\end{table*}

\paragraph{Results.}
Table~\ref{tab:t2_t4_detailed_response} shows that MedPixel achieves
the strongest overall combination of pixel-level prediction and
language generation on both tasks. On T2, MedPixel-7B obtains the best Dice, NSD, METEOR, TGA,
and Head Match results. VisionReasoner is the strongest baseline with
complete reasoning responses, but it remains substantially behind
MedPixel in both segmentation quality and implicit target recovery.
LISA++, PixelLM, and UniPixel mainly produce short templates or
segmentation-oriented outputs, resulting in limited target recovery.
Citrus-V is included for pixel-level comparison, but its per-sample T2
reasoning responses are unavailable.

On T4, MedPixel-7B again achieves the highest Dice, NSD, and METEOR,
together with high target-recovery accuracy.
Citrus-V obtains perfect TGA and Head Match because its short responses
almost always contain the target name. However, its substantially lower
METEOR and segmentation scores indicate that recovering the target
name alone does not constitute a complete and accurately grounded
explanation. VisionReasoner produces much longer responses but remains
behind MedPixel in mask quality, reference similarity, and target
recovery.

Target recovery is generally more difficult on T2 because its query
does not explicitly name the target. The model must first infer the
intended anatomical structure or finding before producing the mask and
reasoning response. T4 instead evaluates the description of an
identified target, leading to higher target-recovery scores and shorter
responses. Overall, MedPixel consistently combines accurate masks with
complete and target-consistent language generation across both
reasoning and explanatory segmentation.

\subsection{Robustness to Bounding-Box Perturbations}

\begin{figure}[t]
\centering
\includegraphics[width=\columnwidth]{
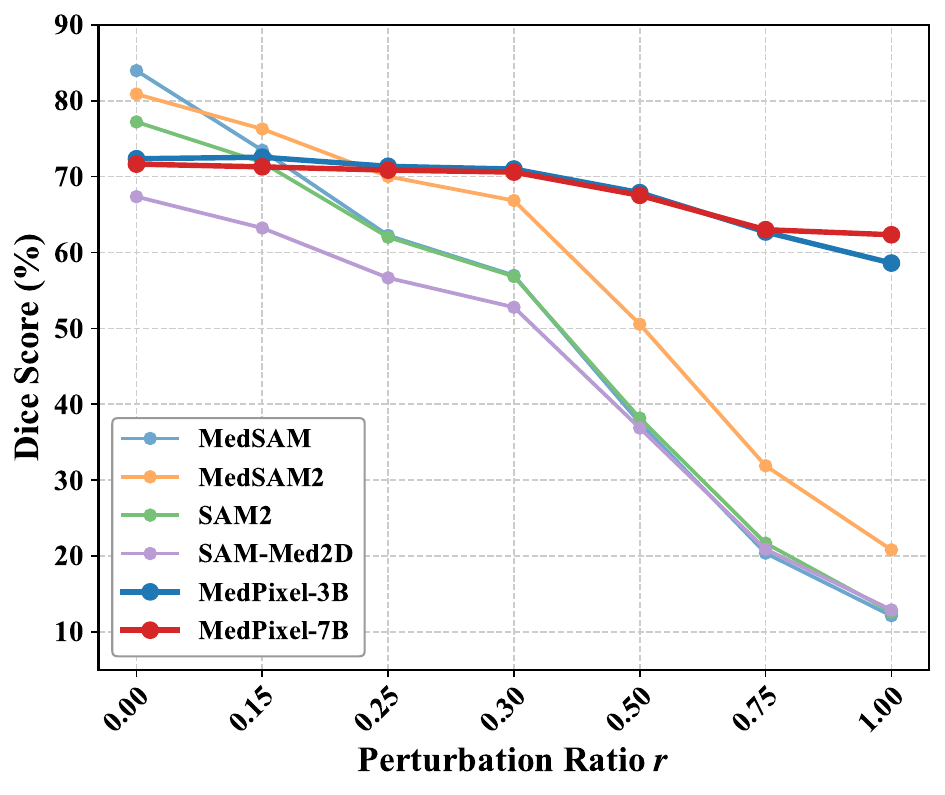
}
\caption{
Dice under joint bounding-box shift-and-scale perturbations.
Larger $r$ indicates stronger center displacement and side variation.
MedPixel degrades more gradually than the SAM-family baselines.
}
\label{fig:box_joint_curve}
\end{figure}

\begin{figure}[!t]
\centering
\includegraphics[width=\columnwidth]{
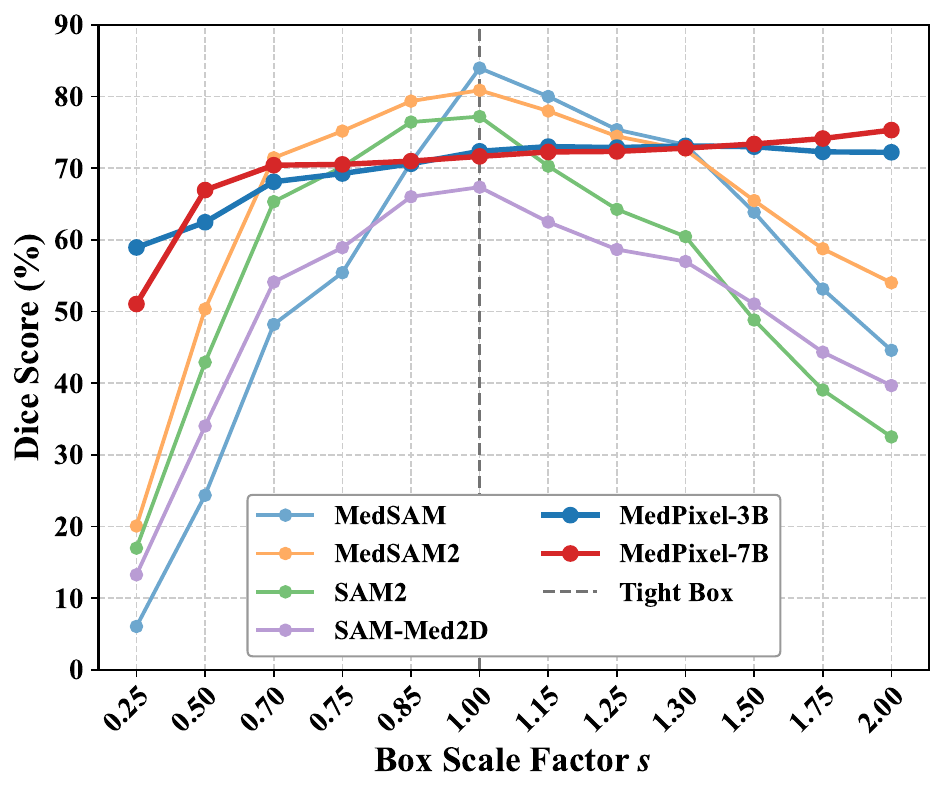
}
\caption{
Dice under scale-only bounding-box perturbations.
The dashed line marks the tight box at $s=1.00$.
MedPixel remains comparatively stable as the box is contracted or
enlarged.
}
\label{fig:box_scale_curve}
\end{figure}

\label{app:spatial_robustness}
\paragraph{Experimental setup.}
We use the same $1{,}000$-sample box test set as in the T3 evaluation,
with a fixed random seed of $42$. All methods receive identically
perturbed boxes and are evaluated against the original ground-truth
masks at their native resolution. We report Dice and normalized surface
Dice (NSD) with a tolerance of five pixels.

\paragraph{Scale-only perturbation.}
To isolate the effect of box size, we keep the original box center
fixed and symmetrically scale its width and height. Let $(c_x,c_y)$,
$w$, and $h$ denote the original box center, width, and height,
respectively. The scaled box is defined as
\begin{equation}
\begin{aligned}
c_x'=c_x,\qquad c_y'=c_y,\qquad w'=sw,\qquad h'=sh.
\end{aligned}
\end{equation}
Here, $s=1.00$ preserves the original tight box, while $s<1.00$
shrinks it and $s>1.00$ enlarges it.

\begin{table*}[t]
\centering
\scriptsize
\setlength{\tabcolsep}{3pt}
\resizebox{\textwidth}{!}{
\begin{tabular}{lcccccccccccc}
\toprule
\textbf{Method}
& $s=0.25$
& $s=0.50$
& $s=0.70$
& $s=0.75$
& $s=0.85$
& $s=1.00$
& $s=1.15$
& $s=1.25$
& $s=1.30$
& $s=1.50$
& $s=1.75$
& $s=2.00$ \\
\midrule
MedSAM & 6.06 & 24.37 & 48.20 & 55.42 & 70.80 & 83.97 & 80.01 & 75.38 & 73.26 & 63.86 & 53.14 & 44.59 \\
MedSAM2 & 20.08 & 50.35 & 71.45 & 75.18 & 79.36 & 80.88 & 77.98 & 74.45 & 72.55 & 65.48 & 58.76 & 54.02 \\
SAM2 & 16.99 & 42.92 & 65.33 & 70.21 & 76.44 & 77.21 & 70.29 & 64.26 & 60.46 & 48.82 & 39.05 & 32.51 \\
SAM-Med2D & 13.27 & 34.03 & 54.12 & 58.92 & 66.02 & 67.36 & 62.49 & 58.66 & 56.98 & 51.06 & 44.33 & 39.67 \\
\midrule
\rowcolor{ourrow}
MedPixel-3B & 58.94 & 62.46 & 68.11 & 69.26 & 70.62 & 72.36 & 73.02 & 72.88 & 73.12 & 73.00 & 72.28 & 72.22 \\
\rowcolor{ourrow}
MedPixel-7B & 51.05 & 66.95 & 70.42 & 70.53 & 70.99 & 71.65 & 72.27 & 72.34 & 72.80 & 73.38 & 74.14 & 75.32 \\
\bottomrule
\end{tabular}}
\caption{
Dice under scale-only bounding-box perturbations.
The box center remains fixed, while its width and height are multiplied
by $s$; $s=1.00$ corresponds to the original tight box.
}
\label{tab:box_scale_robustness}
\end{table*}

As shown in Table~\ref{tab:box_scale_robustness} and
Figure~\ref{fig:box_scale_curve}, the SAM-family methods are strongly
dependent on box scale. Their performance generally peaks near the
original tight box at $s=1.00$ and decreases when the box is either
contracted or enlarged. By contrast, MedPixel exhibits a substantially
flatter performance profile.

MedPixel is particularly stable for enlarged boxes. MedPixel-7B
increases from $71.65$ Dice at $s=1.00$ to $75.32$ at $s=2.00$,
whereas all SAM-family baselines degrade as additional background is
introduced. Stronger degradation is observed under severe contraction,
because the box may no longer contain the complete target.

\paragraph{Joint shift-and-scale perturbation.}
Starting from the tight ground-truth bounding box, we randomly shift its
center in the horizontal and vertical directions, and independently
contract or expand its left, right, top, and bottom sides relative to
the shifted center. The resulting box therefore contains both positional
displacement and asymmetric extent variation.

Let the original box have center $(c_x,c_y)$, width $w$, height $h$,
and center-to-side distances $d_k$, where
$k\in\{\mathrm{l},\mathrm{r},\mathrm{t},\mathrm{b}\}$. We perturb the
box as
\begin{equation}
\begin{aligned}
c_x' &= c_x+\epsilon_x w,
& \epsilon_x &\sim \mathcal{U}(-r,r), \\
c_y' &= c_y+\epsilon_y h,
& \epsilon_y &\sim \mathcal{U}(-r,r), \\
d_k' &= \alpha_k d_k,
& \alpha_k &\sim \mathcal{U}(1-r,1+r).
\end{aligned}
\end{equation}
The four side-scaling factors are sampled independently. The resulting
box is
\begin{equation}
B'=
\left(
c_x'-d_{\mathrm{l}}',
c_y'-d_{\mathrm{t}}',
c_x'+d_{\mathrm{r}}',
c_y'+d_{\mathrm{b}}'
\right).
\end{equation}
Here, $r=0.00$ retains the original tight box, while larger values
produce stronger center displacement and side variation.

\begin{table*}[t]
\centering
\fontsize{7.3}{8.6}\selectfont
\renewcommand{\arraystretch}{1.12}
\begin{adjustbox}{max width=\textwidth}
\begin{tabular}{@{}l*{8}{cc}@{}}
\toprule
\multirow{2}{*}{\textbf{Method}}
& \multicolumn{2}{c}{$r=0.00$}
& \multicolumn{2}{c}{$r=0.15$}
& \multicolumn{2}{c}{$r=0.25$}
& \multicolumn{2}{c}{$r=0.30$}
& \multicolumn{2}{c}{$r=0.50$}
& \multicolumn{2}{c}{$r=0.75$}
& \multicolumn{2}{c}{$r=1.00$}
& \multicolumn{2}{c}{$\Delta_{0\rightarrow1}$} \\
\cmidrule(lr){2-3}\cmidrule(lr){4-5}\cmidrule(lr){6-7}\cmidrule(lr){8-9}
\cmidrule(lr){10-11}\cmidrule(lr){12-13}\cmidrule(lr){14-15}\cmidrule(lr){16-17}
& \textbf{Dice} & \textbf{NSD}
& \textbf{Dice} & \textbf{NSD}
& \textbf{Dice} & \textbf{NSD}
& \textbf{Dice} & \textbf{NSD}
& \textbf{Dice} & \textbf{NSD}
& \textbf{Dice} & \textbf{NSD}
& \textbf{Dice} & \textbf{NSD}
& \textbf{Dice} & \textbf{NSD} \\
\midrule
MedSAM & \textbf{83.97} & \textbf{62.10} & 73.47 & 41.82 & 62.25 & 31.43 & 56.98 & 27.96 & 37.56 & 19.04 & 20.39 & 12.70 & 12.14 & 8.20 & $-71.83$ & $-53.90$ \\
MedSAM2 & 80.88 & 57.14 & \textbf{76.28} & 48.07 & 70.02 & 39.38 & 66.85 & 35.79 & 50.55 & 25.02 & 31.90 & 17.28 & 20.82 & 12.06 & $-60.06$ & $-45.08$ \\
SAM2 & 77.21 & 41.47 & 71.87 & 34.83 & 62.02 & 27.99 & 56.84 & 25.71 & 38.18 & 17.68 & 21.70 & 11.99 & 12.64 & 7.72 & $-64.57$ & $-33.75$ \\
SAM-Med2D & 67.36 & 36.17 & 63.23 & 31.69 & 56.66 & 26.83 & 52.79 & 24.60 & 36.87 & 18.07 & 20.90 & 12.28 & 12.90 & 8.57 & $-54.46$ & $-27.60$ \\
\midrule
\rowcolor{ourrow}
MedPixel-3B & 72.36 & 49.39 & 72.56 & \textbf{49.60} & \textbf{71.36} & 48.45 & \textbf{71.00} & 48.02 & \textbf{67.89} & 45.09 & 62.71 & 41.15 & 58.61 & 38.95 & $-13.75$ & $-10.44$ \\
\rowcolor{ourrow}
MedPixel-7B & 71.65 & 49.79 & 71.29 & 49.46 & 70.86 & \textbf{49.15} & 70.60 & \textbf{48.82} & 67.54 & \textbf{46.66} & \textbf{63.01} & \textbf{43.46} & \textbf{62.34} & \textbf{43.26} & $\mathbf{-9.31}$ & $\mathbf{-6.53}$ \\
\bottomrule
\end{tabular}
\end{adjustbox}
\caption{
Robustness to joint bounding-box shift-and-scale perturbations.
The perturbation ratio $r$ controls both center displacement and
independent variation of the four box sides, with $r=0.00$
corresponding to the original tight box.
}
\label{tab:box_robustness}
\end{table*}

As shown in Table~\ref{tab:box_robustness} and
Figure~\ref{fig:box_joint_curve}, the SAM-family methods perform
strongly with tight boxes but degrade rapidly as the perturbation
increases. Their Dice curves decline sharply once the box is
substantially displaced or asymmetrically resized. In contrast, both
MedPixel variants remain comparatively stable across increasing
perturbation strengths.

At $r=1.00$, MedPixel-7B retains $62.34$ Dice and $43.26$ NSD,
corresponding to decreases of only $9.31$ and $6.53$ points,
respectively. The comparatively flat MedPixel curves indicate that its
predictions are less dependent on precise agreement between the input
box and the target extent.

\paragraph{Qualitative analysis.}
We further examine representative endoscopic-polyp and chest-X-ray examples under both perturbation settings. In the joint-perturbation visualizations, green, magenta, and yellow indicate the ground-truth contour, predicted contour, and perturbed box, respectively, and $D$ denotes the Dice score of the displayed prediction.

\begin{figure*}[p]
\centering
\includegraphics[
    width=\textwidth,
    height=0.84\textheight,
    keepaspectratio
]{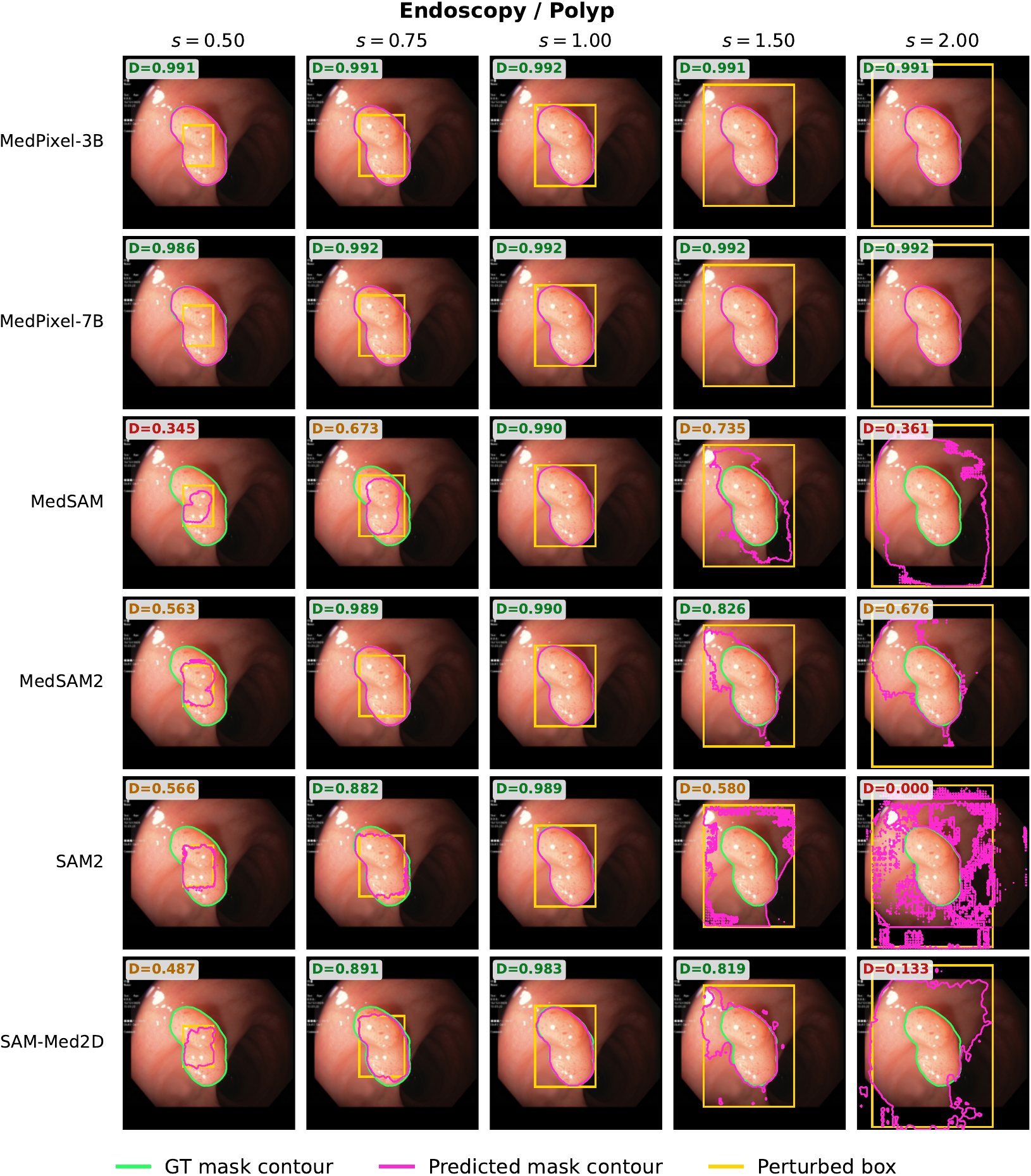}
\caption{
Qualitative robustness to scale-only perturbations for an endoscopic
polyp. MedPixel preserves an accurate target contour across both
contracted and enlarged boxes, while the SAM-family methods are more
sensitive to deviations from the original tight box.
}
\label{fig:scale_perturbation_polyp}
\end{figure*}

\begin{figure*}[p]
\centering
\includegraphics[
    width=\textwidth,
    height=0.84\textheight,
    keepaspectratio
]{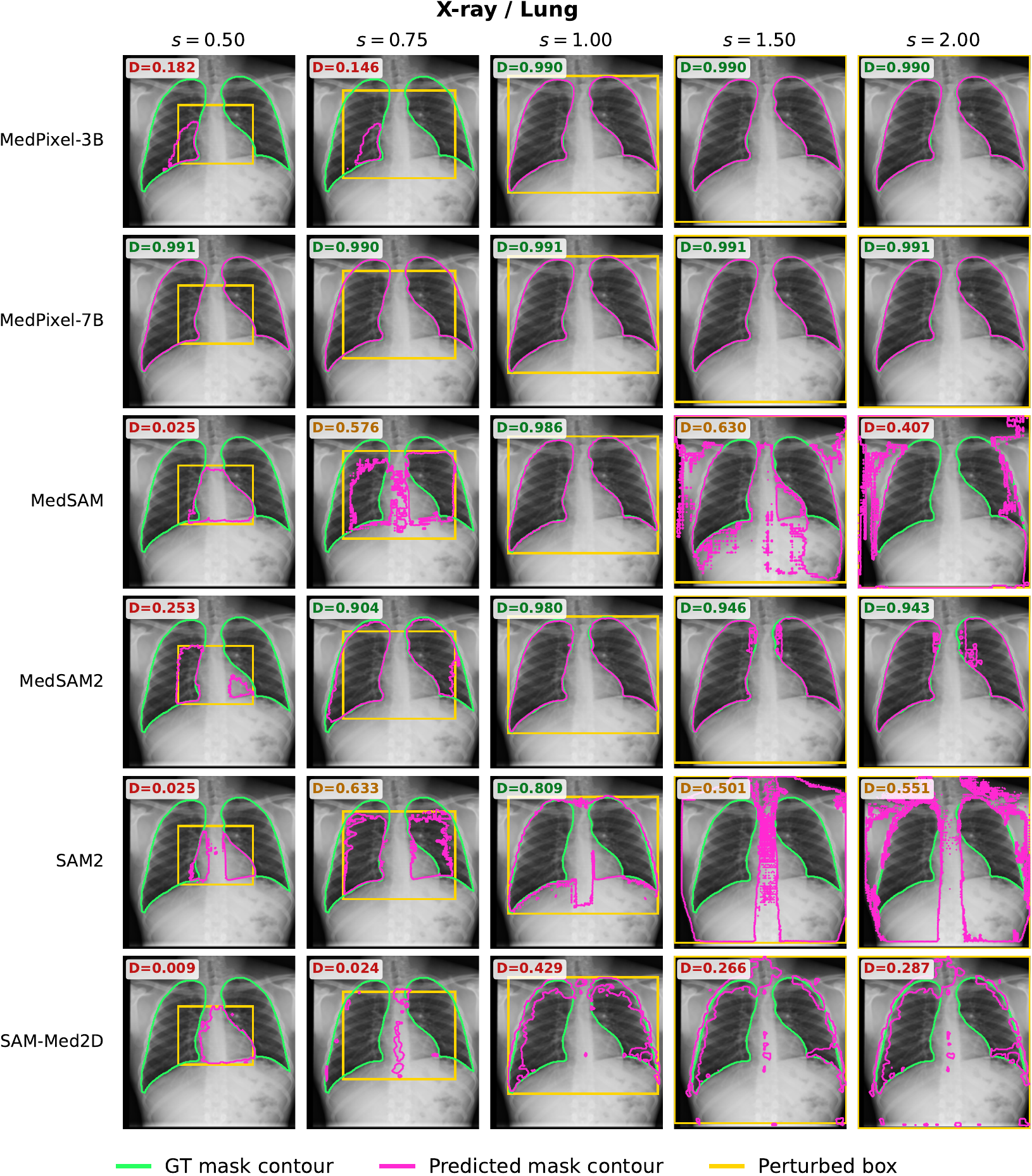}
\caption{
Qualitative robustness to scale-only perturbations for the lungs in a
chest X-ray. Severe contraction removes a substantial portion of the
target and is therefore more challenging than box enlargement.
MedPixel-7B remains stable across the evaluated scales, whereas
MedPixel-3B is more sensitive when the box covers only a small central
part of the lungs.
}
\label{fig:scale_perturbation_lung}
\end{figure*}

\noindent\textit{Scale-only examples.}
Figures~\ref{fig:scale_perturbation_polyp}
and~\ref{fig:scale_perturbation_lung} visualize the effect of changing
the box size while preserving its center. The two examples illustrate
the different effects of box contraction and enlargement across target
scales.

The scale-only examples explain the asymmetric behavior observed in
Table~\ref{tab:box_scale_robustness}. Enlarged boxes preserve the
complete target and mainly introduce additional background, which
MedPixel can largely suppress. In contrast, strongly contracted boxes
may remove essential target regions and provide insufficient spatial
evidence. This effect is especially visible for the large bilateral
lung target, where MedPixel-7B is better able to recover the complete
anatomy from an incomplete box. For the smaller and visually distinctive
polyp, both MedPixel variants remain stable over a broad range of box
scales.

As a geometric reference, using the tight box directly as the
foreground mask yields $66.18$ Dice and $15.44$ NSD, confirming that
MedPixel predicts target boundaries rather than simply reproducing the
box. Together, the quantitative and qualitative results show that
MedPixel can use image and target evidence to remain stable under
inaccurate spatial prompts, with degradation occurring primarily when
severe perturbations remove a substantial portion of the target.

\begin{figure*}[p]
\centering
\includegraphics[
    width=\textwidth,
    height=0.84\textheight,
    keepaspectratio
]{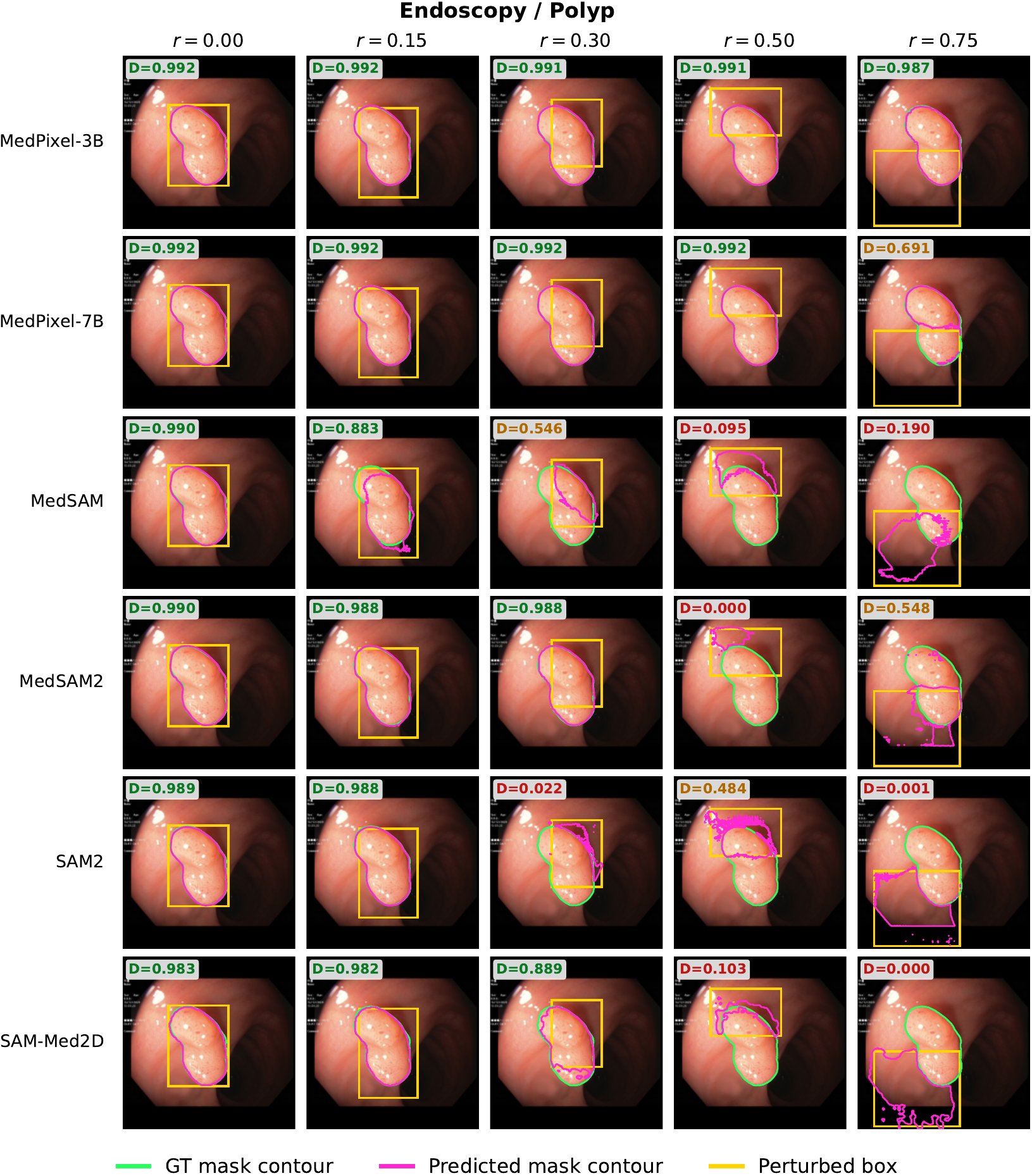}
\caption{
Qualitative robustness to joint shift-and-scale perturbations for an
endoscopic polyp. The perturbation strength increases from left to
right. MedPixel remains closely aligned with the target under moderate
box displacement and asymmetric resizing, whereas the compared
SAM-family methods become increasingly sensitive to the perturbed box.
}
\label{fig:joint_perturbation_polyp}
\end{figure*}

\begin{figure*}[p]
\centering
\includegraphics[
    width=\textwidth,
    height=0.84\textheight,
    keepaspectratio
]{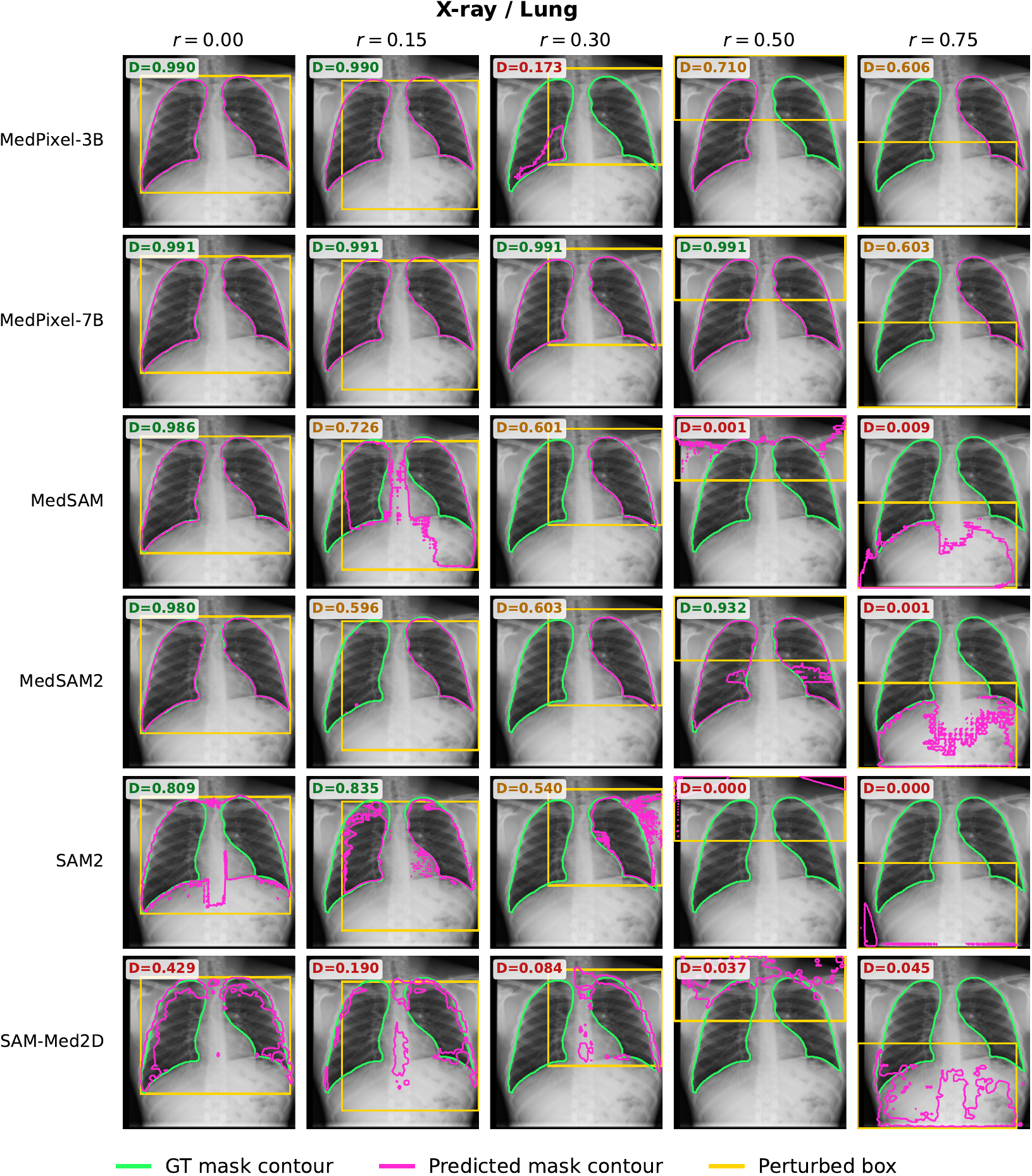}
\caption{
Qualitative robustness to joint shift-and-scale perturbations for the
lungs in a chest X-ray. The lung example is more challenging because a
displaced box may exclude a large portion of the bilateral target.
MedPixel-7B remains accurate under moderate perturbations, while severe
target exclusion eventually causes degradation.
}
\label{fig:joint_perturbation_lung}
\end{figure*}

\noindent\textit{Joint-perturbation examples.}
Figures~\ref{fig:joint_perturbation_polyp}
and~\ref{fig:joint_perturbation_lung} provide representative
comparisons for a localized endoscopic lesion and a large bilateral
anatomical structure. 

The qualitative examples support the aggregate trend in
Table~\ref{tab:box_robustness}. For the localized polyp, MedPixel
continues to recover the target contour even when the perturbed box is
visibly displaced or asymmetrically resized. The lung example reveals
a more difficult failure mode: when the box excludes a substantial
portion of the bilateral anatomy, the prediction may become incomplete.
Nevertheless, MedPixel-7B remains stable over a wider perturbation range,
whereas the SAM-family methods frequently fragment, expand into
background regions, or collapse to an incorrect structure.

\section{Qualitative Results}
\label{app:qualitative_results}

\begin{figure*}[p]
\centering
\includegraphics[
    page=1,
    width=\textwidth,
    height=0.88\textheight,
    keepaspectratio
]{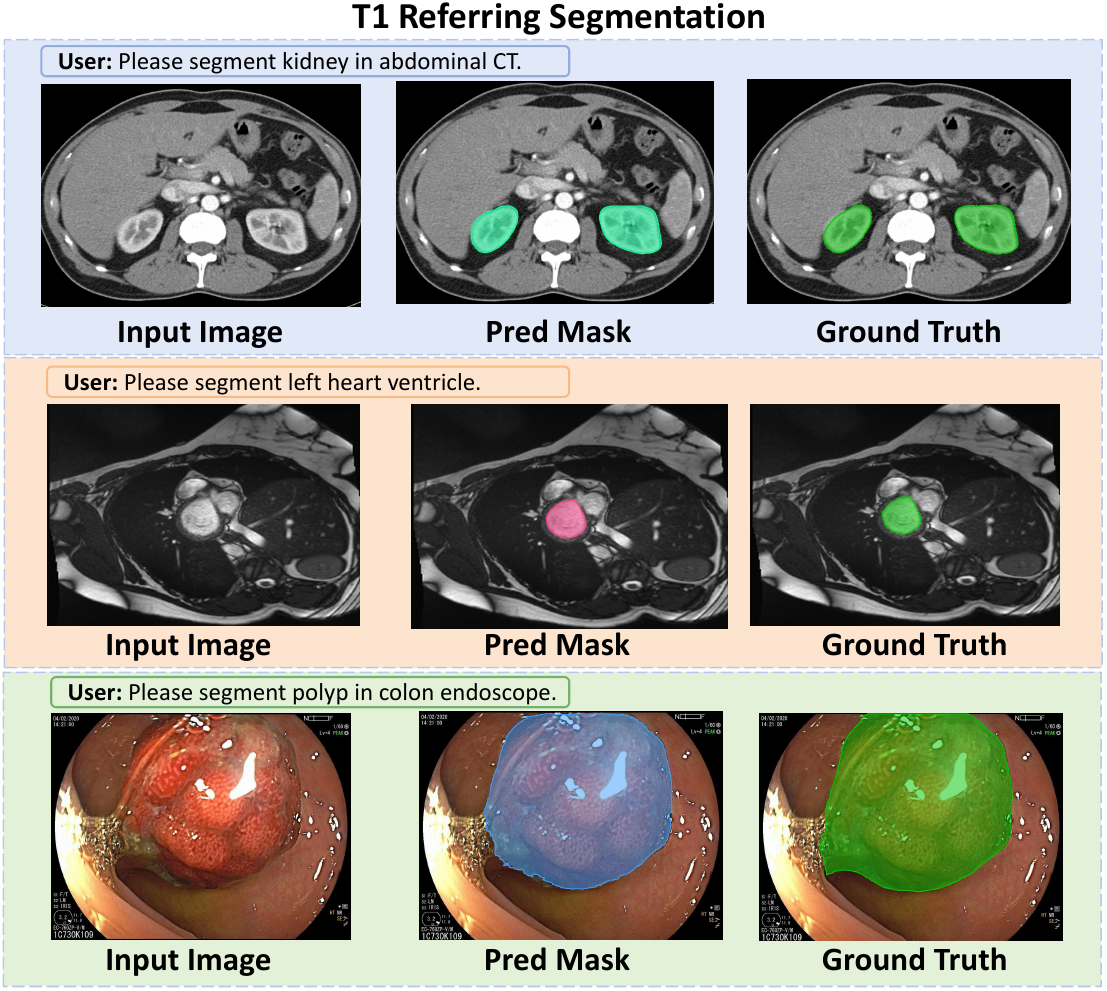}
\caption{\textbf{Qualitative results for T1 referring segmentation.}
Given an explicitly specified target, MedPixel directly predicts the
corresponding mask. The examples cover kidney segmentation in
abdominal CT, left-ventricle segmentation in cardiac MRI, and polyp
segmentation in colonoscopy.}
\label{fig:qual_t1}
\end{figure*}

\begin{figure*}[p]
\centering
\includegraphics[
    page=2,
    width=\textwidth,
    height=0.88\textheight,
    keepaspectratio
]{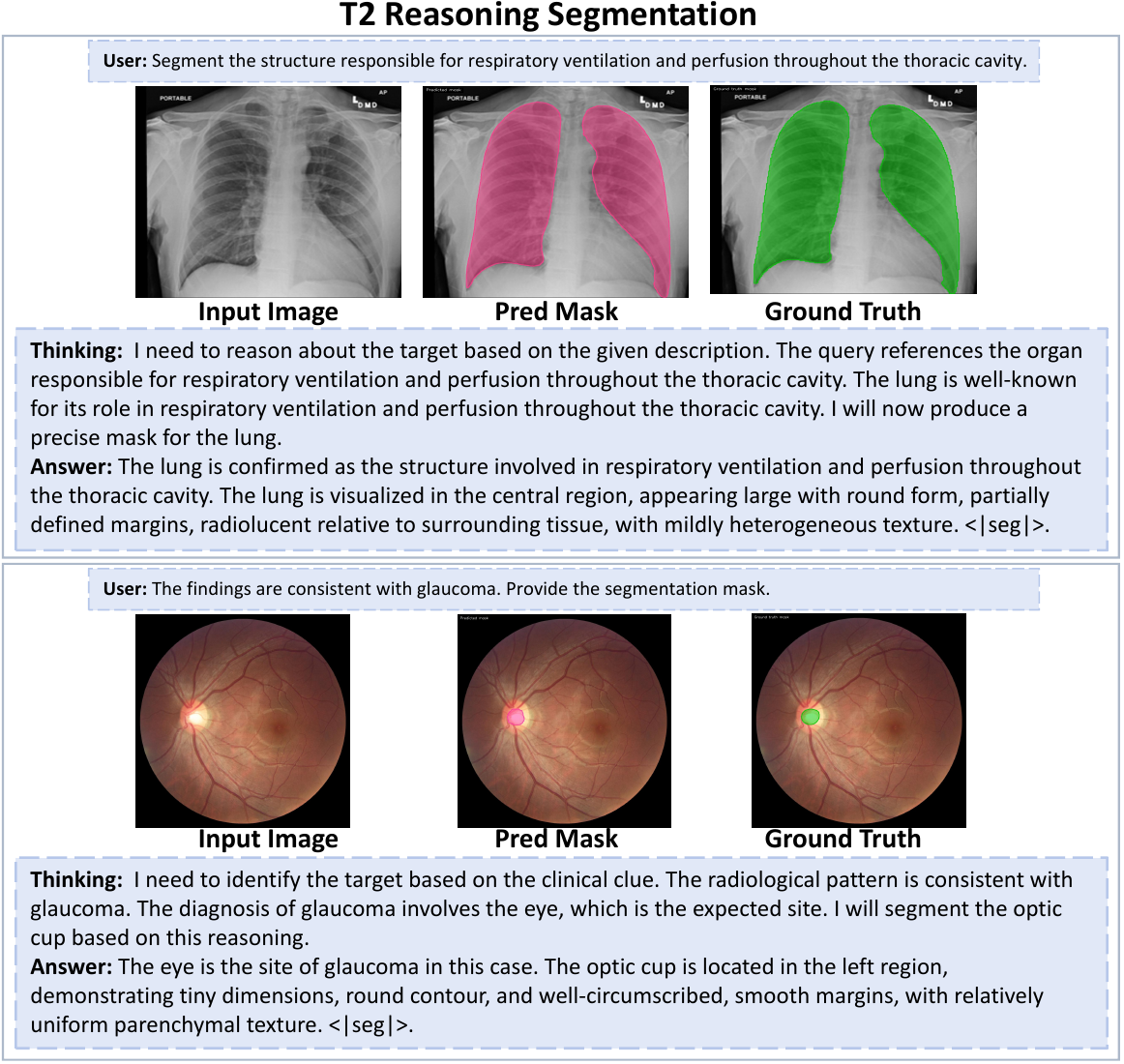}
\caption{\textbf{Qualitative results for T2 reasoning segmentation.}
MedPixel infers the intended target from functional or clinical clues
before producing the grounded response and segmentation mask. The
examples identify the lungs from their physiological function and the
optic cup from a glaucoma-related diagnostic clue.}
\label{fig:qual_t2}
\end{figure*}

\begin{figure*}[p]
\centering
\includegraphics[
    page=3,
    width=\textwidth,
    height=0.88\textheight,
    keepaspectratio
]{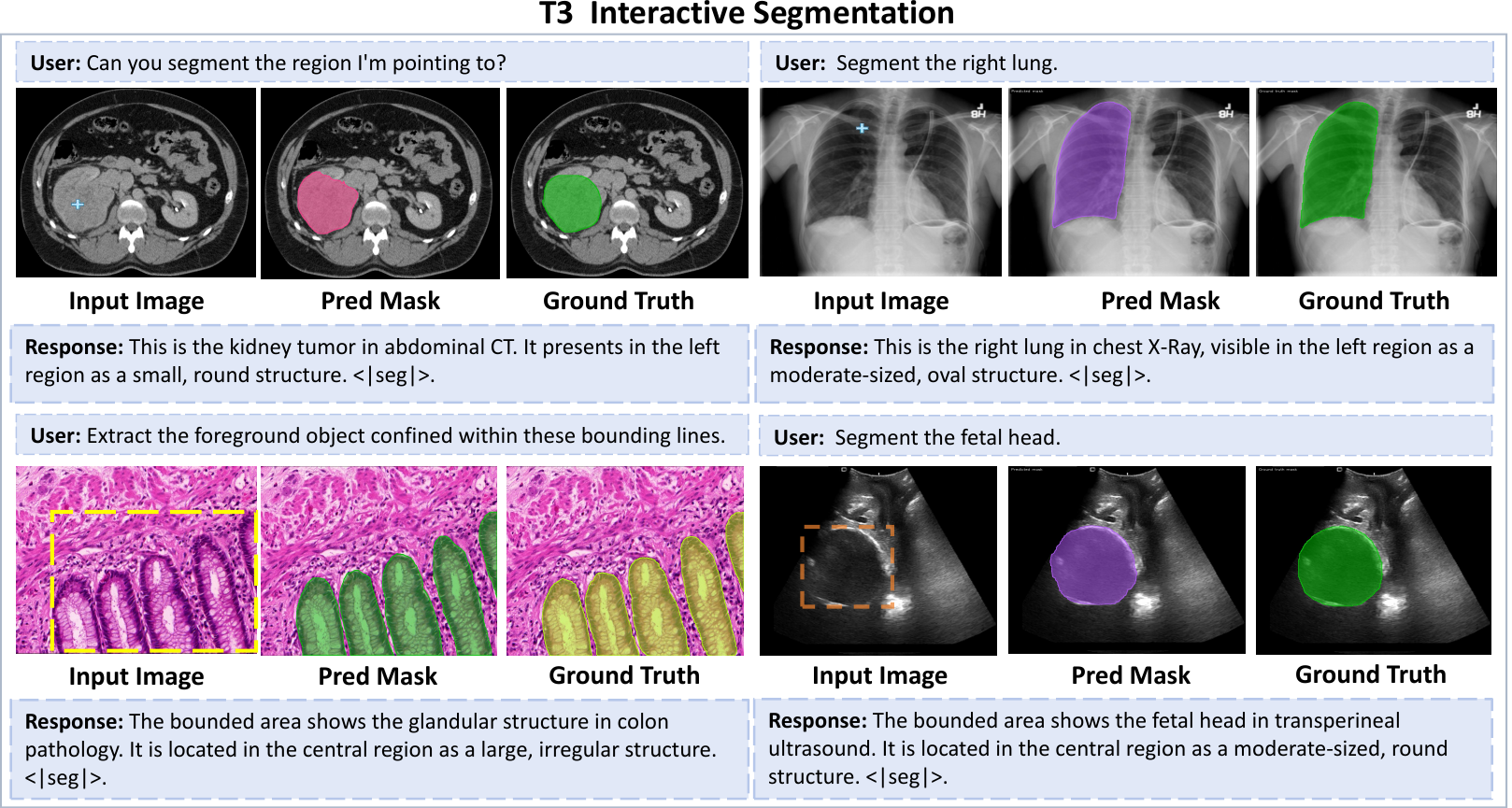}
\caption{\textbf{Qualitative results for T3 interactive segmentation.}
MedPixel identifies and segments targets indicated by point or
bounding-box prompts. The examples cover abdominal CT, chest
radiography, colon pathology, and transperineal ultrasound. The
generic textual instruction specifies the requested operation, while
the spatial prompt determines the target region.}
\label{fig:qual_t3}
\end{figure*}

\begin{figure*}[p]
\centering
\includegraphics[
    page=4,
    width=\textwidth,
    height=0.88\textheight,
    keepaspectratio
]{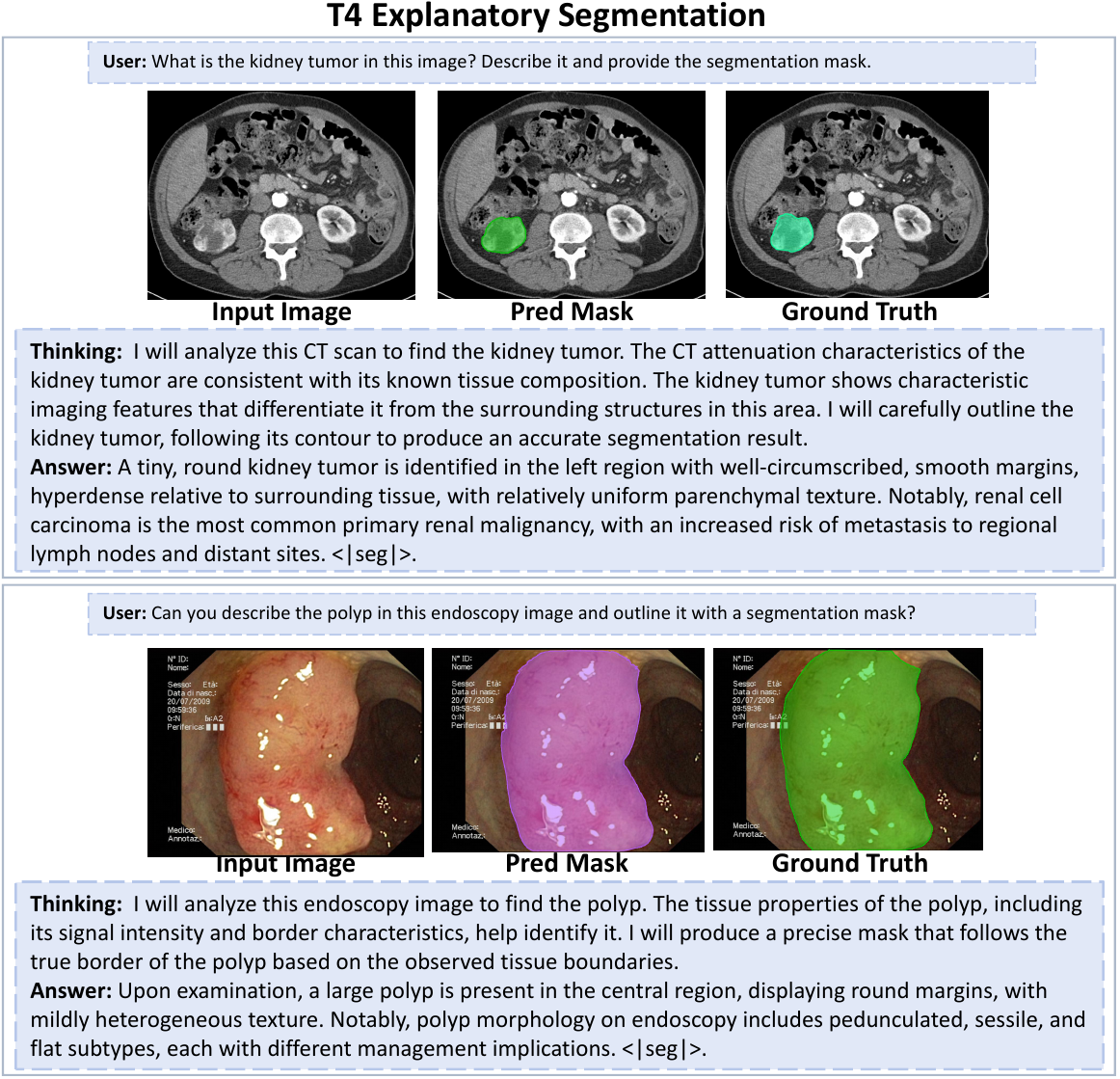}
\caption{\textbf{Qualitative results for T4 explanatory segmentation.}
MedPixel jointly generates a visual explanation and a segmentation
mask. The responses identify and describe the location, morphology,
boundary, and appearance of a kidney tumor in abdominal CT and a
polyp in endoscopy.}
\label{fig:qual_t4}
\end{figure*}

\begin{figure*}[p]
\centering
\includegraphics[
    page=5,
    width=\textwidth,
    height=0.88\textheight,
    keepaspectratio
]{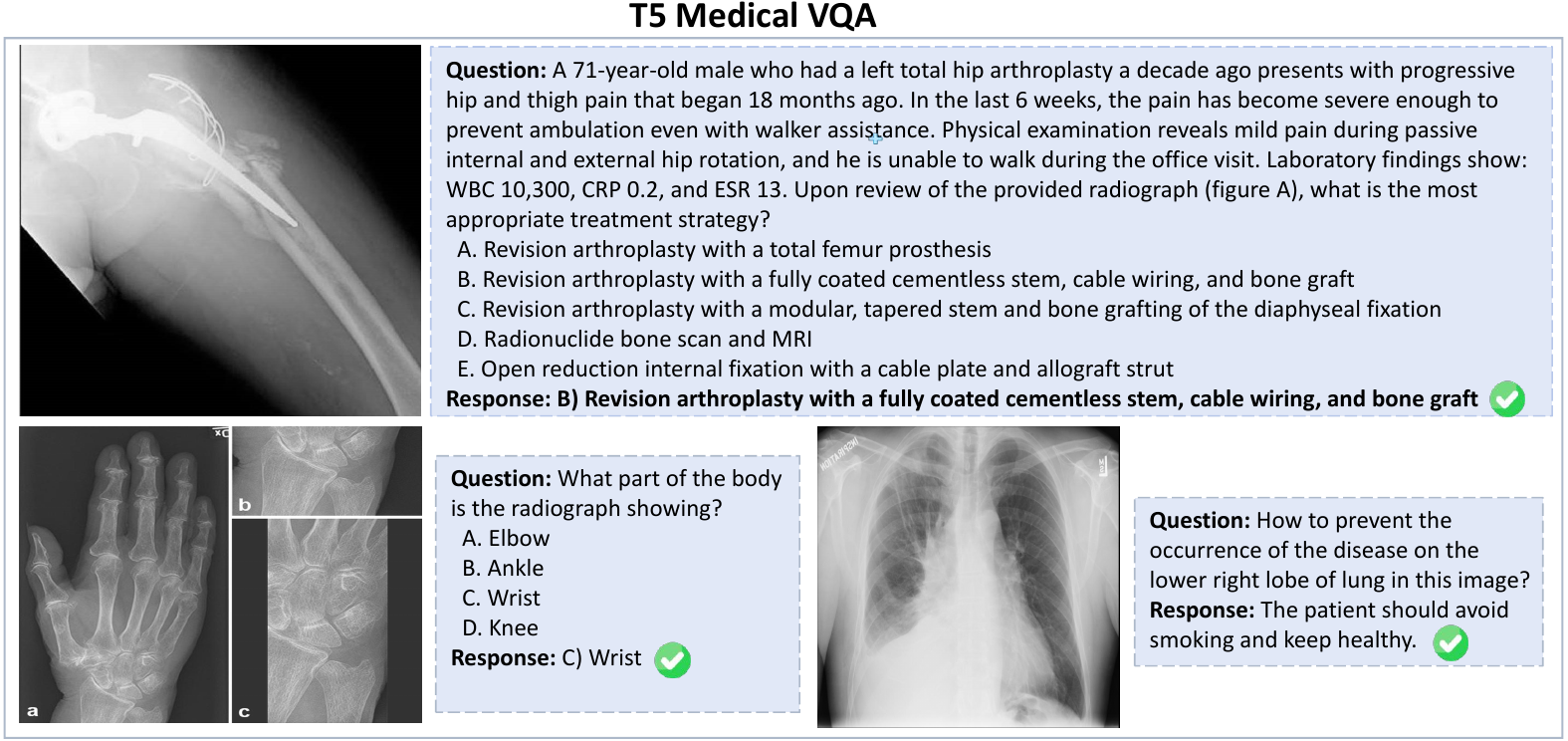}
\caption{\textbf{Qualitative results for T5 medical VQA.}
MedPixel retains general medical question-answering capability without
requiring pixel-level output. The examples cover treatment-strategy
selection, preventive guidance based on a chest radiograph, and
anatomical recognition from radiographs.}
\label{fig:qual_t5}
\end{figure*}

Figures~\ref{fig:qual_t1}--\ref{fig:qual_t5} present representative
outputs across the five MedPixel task types. For T1--T4, each example
shows the user prompt, input image, predicted mask, and ground-truth
mask. T2 and T4 additionally display the generated reasoning and final
grounded response, while T3 demonstrates interaction through point or
bounding-box prompts. T5 presents medical question-answering examples
that do not require pixel-level prediction.

Across the five task types, MedPixel adapts its output to the requested
interaction form. T1 directly grounds an explicitly named target,
whereas T2 infers the intended structure from functional or clinical
evidence before localization. T3 follows point- and box-based spatial
prompts, and T4 couples visual explanation with mask prediction. T5
further demonstrates that the unified model preserves medical
question-answering capability when no pixel-level output is required.

\subsection{Licenses}
\label{app:licenses}

MedPLG-440K is derived from datasets aggregated by
BiomedParse~\cite{biomedparse}. These source datasets are distributed
under different licenses and usage conditions, and downstream users
remain responsible for complying with the terms specified by the
original data providers. We do not redistribute the underlying medical
images. MedPLG-440K and MedPixel are intended for research use only and
have not been validated for clinical diagnosis or treatment decisions.

\end{document}